\documentclass[10pt,twocolumn,letterpaper]{article}

\usepackage[pagenumbers]{cvpr}              

\usepackage{multirow}
\usepackage{comment}

\usepackage{graphicx}
\usepackage{subcaption}
\usepackage{tikz}
\usepackage{amsmath}
\usepackage{xcolor}
\usepackage{pgfplots}
\pgfplotsset{compat=1.17}
\usepackage{amsmath}
\usepackage{pgfplots}
\usepackage{pgfplotstable}

\definecolor{cvprblue}{rgb}{0.21,0.49,0.74}
\usepackage[pagebackref,breaklinks,colorlinks,allcolors=cvprblue]{hyperref}

\def\paperID{9234} 
\def\confName{CVPR}
\def\confYear{2025}

\title{Enhancing Facial Privacy Protection via Weakening Diffusion Purification}

\author{Ali Salar$^{1}$ \quad Qing Liu$^{1}$ \quad Yingli Tian$^{2}$ \quad Guoying Zhao$^{1*}$ \vspace{0.4em} \\
{\small $^1$Center for Machine Vision and Signal Analysis (CMVS), University of Oulu, Finland}\\ \quad
{\tt\small \{ali.salar, qing.liu, guoying.zhao\}@oulu.fi} \\
{\small $^2$City University of New York (CUNY)} \quad
{\tt\small ytian@ccny.cuny.edu}}

\begin{document}

\twocolumn[{
    \maketitle
    \begin{center}
        \centering
        \captionsetup{type=figure}
        \resizebox{0.9\textwidth}{!}
        {\begin{tikzpicture}
            \node[] at (2.04, 2.7) {\textbf{\fontsize{6.5pt}{8.5pt}\selectfont Noise-based}};
            \node[] at (6.163, 2.7) {\textbf{\fontsize{6.5pt}{8.5pt}\selectfont Makeup-based}};
            \node[] at (11.29, 2.7) {\textbf{\fontsize{6.5pt}{8.5pt}\selectfont Diffusion-based}};
            \node[] at (0, 0){
                \begin{subfigure}[b]{0.11\textwidth}
                    \includegraphics[width=\linewidth]{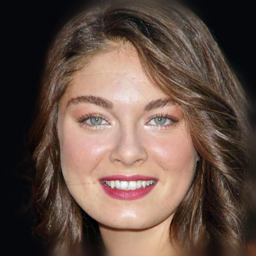}
                    \centering
                    \vspace{-15pt}
                    \caption*{\colorbox{black}{\parbox{\dimexpr\linewidth-2\fboxsep}{\centering\textbf{\fontsize{6pt}{8pt}\selectfont\textcolor[rgb]{1, 0.75, 0}{48.38}}}}}
                    \vspace{2pt}
                    \includegraphics[width=\linewidth]{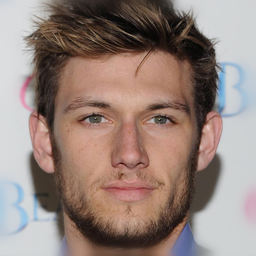}
                    \centering
                    \vspace{-15pt}
                    \caption*{\colorbox{black}{\parbox{\dimexpr\linewidth-2\fboxsep}{\centering\textbf{\fontsize{6pt}{8pt}\selectfont\textcolor[rgb]{1, 0.75, 0}{38.58}}}}}
                    \vspace{2pt}
                    \fontsize{6pt}{8pt}\selectfont\textcolor{black}{Original Images}
                \end{subfigure}
            };
            \node[draw, thick, color=black!80, fill=orange!30, inner sep=0.06cm] at (2.04, 0) {
                \begin{subfigure}[b]{0.11\textwidth}
                    \includegraphics[width=\linewidth]{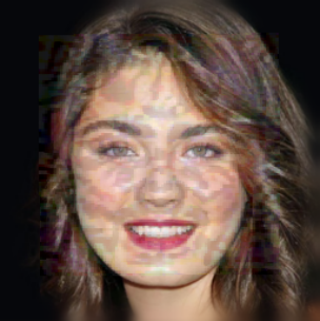}
                    \centering
                    \vspace{-15pt}
                    \caption*{\colorbox{black}{\parbox{\dimexpr\linewidth-2\fboxsep}{\centering\textbf{\fontsize{6pt}{8pt}\selectfont\textcolor[rgb]{1, 0.75, 0}{66.25}}}}}
                    \vspace{2pt}
                    \includegraphics[width=\linewidth]{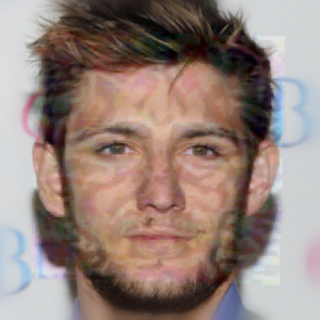}
                    \centering
                    \vspace{-15pt}
                    \caption*{\colorbox{black}{\parbox{\dimexpr\linewidth-2\fboxsep}{\centering\textbf{\fontsize{6pt}{8pt}\selectfont\textcolor[rgb]{1, 0.75, 0}{46.06}}}}}
                    \vspace{2pt}
                    \fontsize{6pt}{8pt}\selectfont\textcolor{black}{(a) TIP-IM \cite{TIP-IM_2021_CVPR}}
                \end{subfigure}
            };
            \node[draw, thick, color=black!80, fill=green!30, inner sep=0.06cm] at (6.163, 0) {
                \begin{subfigure}[b]{0.11\textwidth}
                    \includegraphics[width=\linewidth]{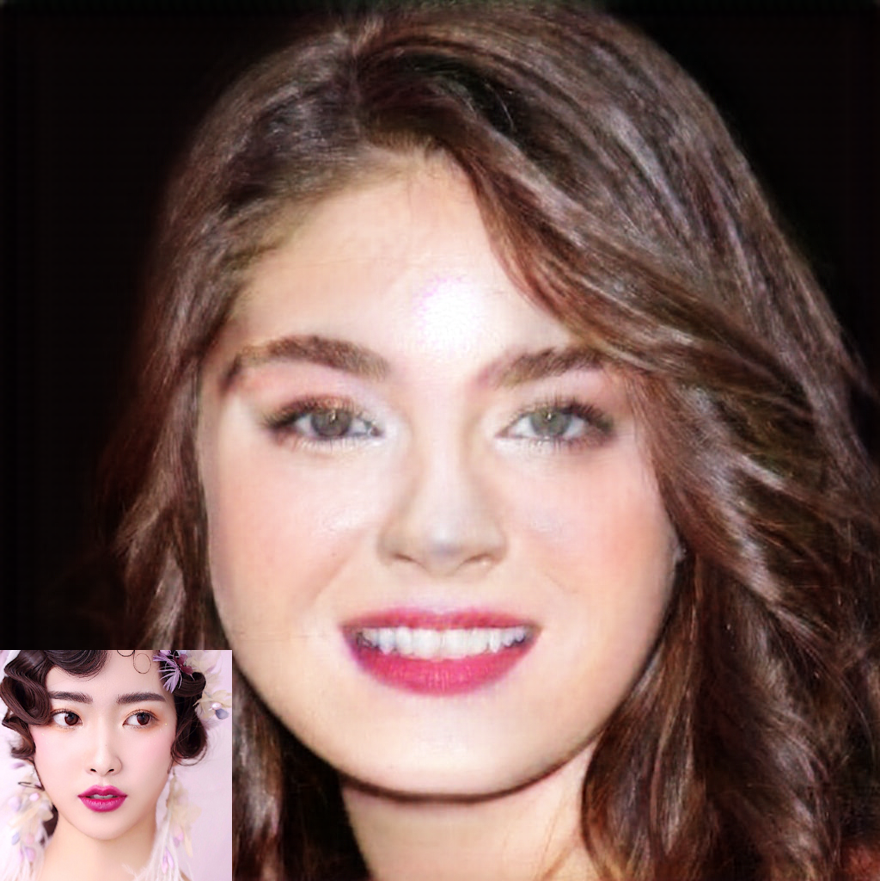}
                    \centering
                    \vspace{-15pt}
                    \caption*{\colorbox{black}{\parbox{\dimexpr\linewidth-2\fboxsep}{\centering\textbf{\fontsize{6pt}{8pt}\selectfont\textcolor[rgb]{1, 0.75, 0}{58.21}}}}}
                    \vspace{2pt}
                    \includegraphics[width=\linewidth]{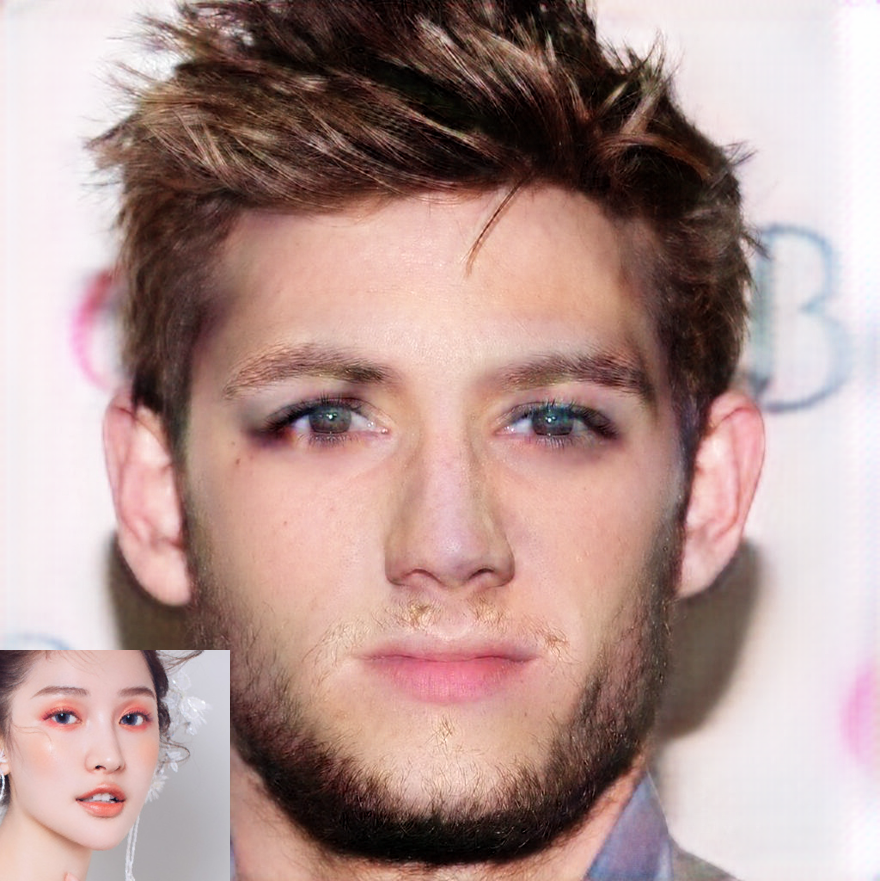}
                    \centering
                    \vspace{-15pt}
                    \caption*{\colorbox{black}{\parbox{\dimexpr\linewidth-2\fboxsep}{\centering\textbf{\fontsize{6pt}{8pt}\selectfont\textcolor[rgb]{1, 0.75, 0}{58.36}}}}}
                    \vspace{2pt}
                    \fontsize{6pt}{8pt}\selectfont\textcolor{black}{(b) AMT-GAN \cite{AMT-GAN_2022_CVPR}}
                \end{subfigure}
                \begin{subfigure}[b]{0.11\textwidth}
                    \includegraphics[width=\linewidth]{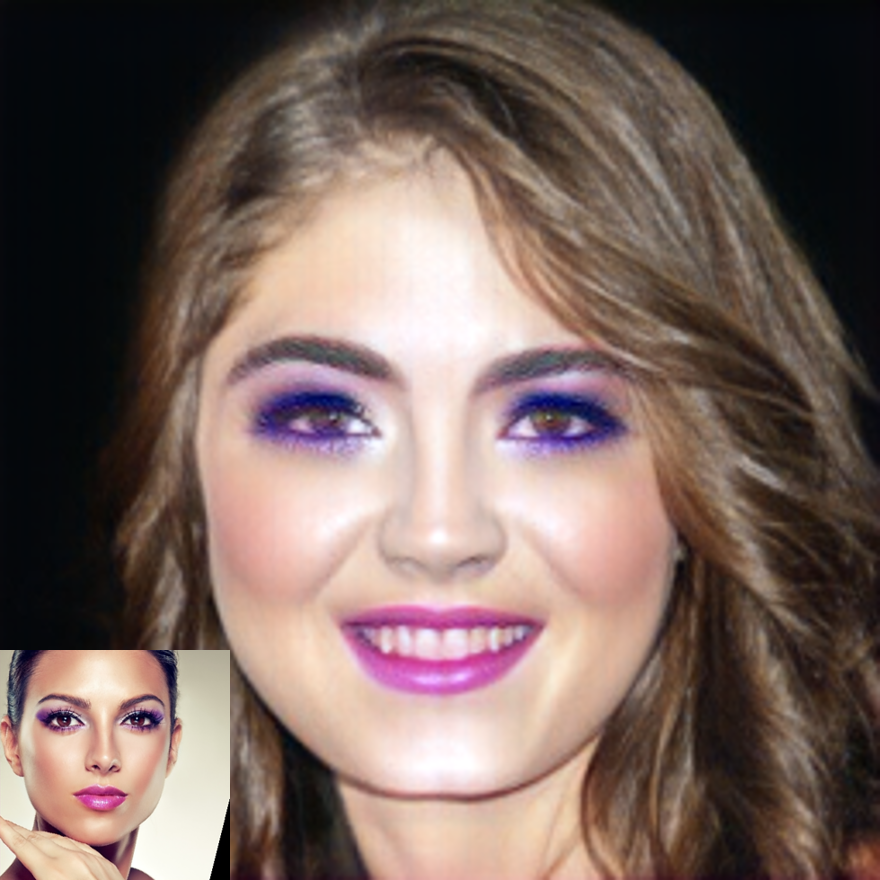}
                    \centering
                    \vspace{-15pt}
                    \caption*{\colorbox{black}{\parbox{\dimexpr\linewidth-2\fboxsep}{\centering\textbf{\fontsize{6pt}{8pt}\selectfont\textcolor[rgb]{1, 0.75, 0}{73.74}}}}}
                    \vspace{2pt}
                    \includegraphics[width=\linewidth]{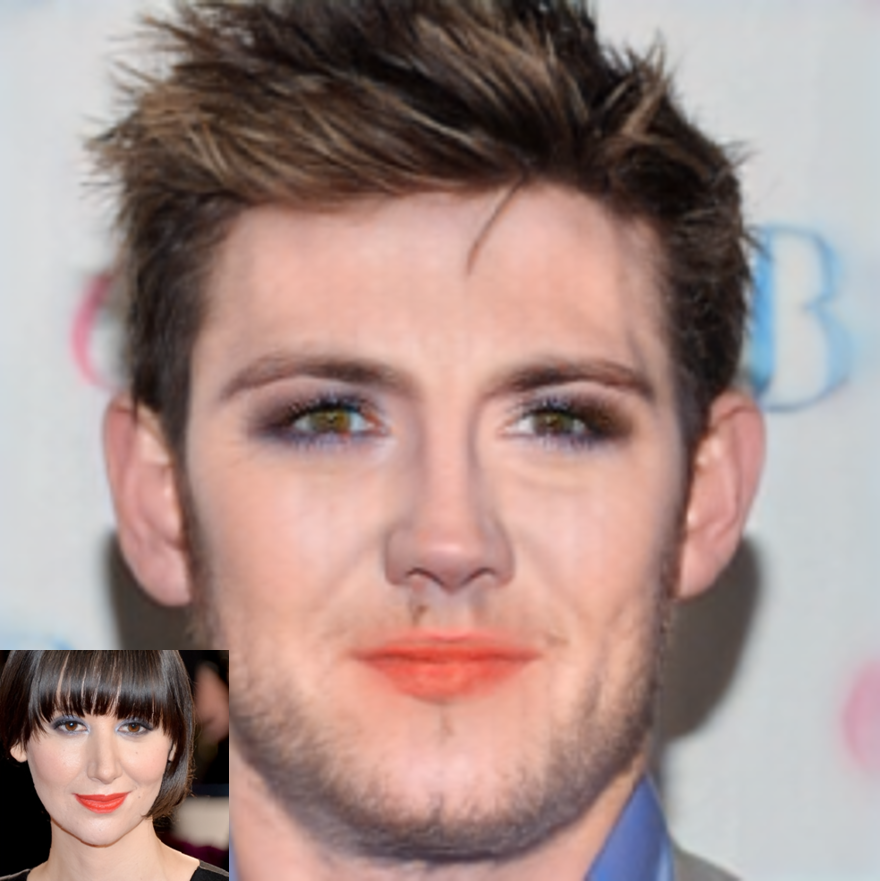}
                    \centering
                    \vspace{-15pt}
                    \caption*{\colorbox{black}{\parbox{\dimexpr\linewidth-2\fboxsep}{\centering\textbf{\fontsize{6pt}{8pt}\selectfont\textcolor[rgb]{1, 0.75, 0}{64.92}}}}}
                    \vspace{2pt}
                    \fontsize{6pt}{8pt}\selectfont\textcolor{black}{(c) DiffAM \cite{DiffAM_2024_CVPR}}
                \end{subfigure}
                \begin{subfigure}[b]{0.11\textwidth}
                    \includegraphics[width=\linewidth]{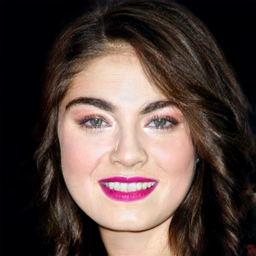}
                    \centering
                    \vspace{-15pt}
                    \caption*{\colorbox{black}{\parbox{\dimexpr\linewidth-2\fboxsep}{\centering\textbf{\fontsize{6pt}{8pt}\selectfont\textcolor[rgb]{1, 0.75, 0}{71.18}}}}}
                    \vspace{2pt}
                    \includegraphics[width=\linewidth]{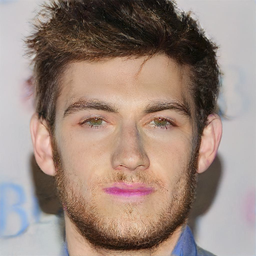}
                    \centering
                    \vspace{-15pt}
                    \caption*{\colorbox{black}{\parbox{\dimexpr\linewidth-2\fboxsep}{\centering\textbf{\fontsize{6pt}{8pt}\selectfont\textcolor[rgb]{1, 0.75, 0}{63.56}}}}}
                    \vspace{2pt}
                    \fontsize{6pt}{8pt}\selectfont\textcolor{black}{(d) CLIP2Protect \cite{Clip2protect_2023_CVPR}}
                \end{subfigure}
                };
            \node[draw, thick,color=black!80, fill=cyan!30, inner sep=0.06cm] at (11.29, 0) {
                \begin{subfigure}[b]{0.11\textwidth}
                    \includegraphics[width=\linewidth]{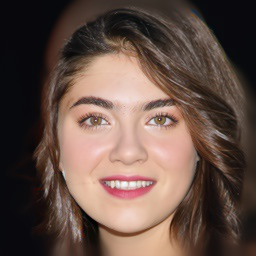}
                    \centering
                    \vspace{-15pt}
                    \caption*{\colorbox{black}{\parbox{\dimexpr\linewidth-2\fboxsep}{\centering\textbf{\fontsize{6pt}{8pt}\selectfont\textcolor[rgb]{1, 0.75, 0}{73.71}}}}}
                    \vspace{2pt}
                    \includegraphics[width=\linewidth]{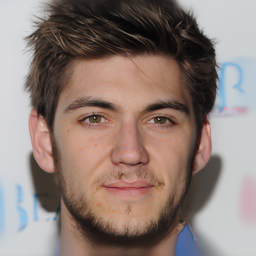}
                    \centering
                    \vspace{-15pt}
                    \caption*{\colorbox{black}{\parbox{\dimexpr\linewidth-2\fboxsep}{\centering\textbf{\fontsize{6pt}{8pt}\selectfont\textcolor[rgb]{1, 0.75, 0}{59.15}}}}}
                    \vspace{2pt}
                    \fontsize{6pt}{8pt}\selectfont\textcolor{black}{(e) DiffProtect \cite{Diffprotect_2023_arXiv}}
                \end{subfigure}
                \begin{subfigure}[b]{0.11\textwidth}
                    \includegraphics[width=\linewidth]{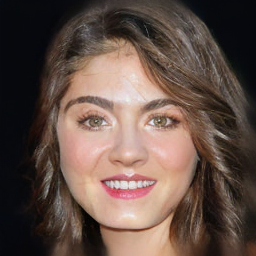}
                    \centering
                    \vspace{-15pt}
                    \caption*{\colorbox{black}{\parbox{\dimexpr\linewidth-2\fboxsep}{\centering\textbf{\fontsize{6pt}{8pt}\selectfont\textcolor[rgb]{1, 0.75, 0}{83.08}}}}}
                    \vspace{2pt}
                    \includegraphics[width=\linewidth]{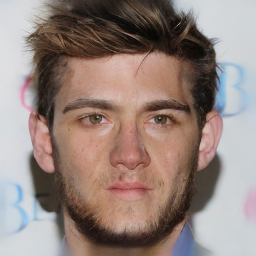}
                    \centering
                    \vspace{-15pt}
                    \caption*{\colorbox{black}{\parbox{\dimexpr\linewidth-2\fboxsep}{\centering\textbf{\fontsize{6pt}{8pt}\selectfont\textcolor[rgb]{1, 0.75, 0}{73.88}}}}}
                    \vspace{2pt}
                    \fontsize{6pt}{8pt}\selectfont\textcolor{black}{Ours}
                \end{subfigure}
            };
        \end{tikzpicture}}
        \begin{minipage}[b]{0.9\textwidth}
            \caption{Protected images generated by various facial privacy protection methods. Below each image, the numbers indicate the verification confidence from the Face++ API; higher values suggest a stronger match with the target image shown on the right side. For (b) and (c), the reference images used for makeup transfer are displayed near the corresponding protected images. In (d), the makeup prompt is “big eyebrows with pink eyeshadows.”}
            \label{fig_1}
        \end{minipage}
        \begin{minipage}[b]{0.09\linewidth}
            \centering
            \includegraphics[width=0.8\linewidth]{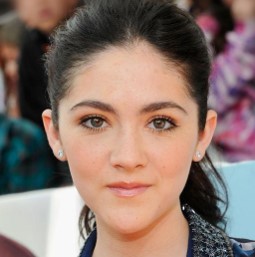}
        \end{minipage}
    \end{center}%
}]

\begin{abstract}
\let\thefootnote\relax\footnotetext{* indicates the corresponding author}
The rapid growth of social media has led to the widespread sharing of individual portrait images, which pose serious privacy risks due to the capabilities of automatic face recognition (AFR) systems for mass surveillance. Hence, protecting facial privacy against unauthorized AFR systems is essential. Inspired by the generation capability of the emerging diffusion models, recent methods employ diffusion models to generate adversarial face images for privacy protection. However, they suffer from the diffusion purification effect, leading to a low protection success rate (PSR). In this paper, we first propose learning unconditional embeddings to increase the learning capacity for adversarial modifications and then use them to guide the modification of the adversarial latent code to weaken the diffusion purification effect. Moreover, we integrate an identity-preserving structure to maintain structural consistency between the original and generated images, allowing human observers to recognize the generated image as having the same identity as the original. Extensive experiments conducted on two public datasets, i.e., CelebA-HQ and LADN, demonstrate the superiority of our approach. The protected faces generated by our method outperform those produced by existing facial privacy protection approaches in terms of transferability and natural appearance. The code is available at 
\href{https://github.com/parham1998/Facial-Privacy-Protection}{https://github.com/parham1998/Facial-Privacy-Protection}
\end{abstract}    
\section{Introduction}
\label{sec:intro}
The advancement of facial recognition (FR) technology, particularly those based on deep neural networks \cite{face_recognition_2021_Neurocomputing, Mobileface_2019_CVPR}, has led to widespread applications in various fields, such as biometrics \cite{biometric_recognition_2024_ACM}, surveillance \cite{surveillance_2022_book}, etc. Despite the benefits, FR technology could be misused by unauthorized organizations to e.g. track users' interactions and activities, which poses serious risks to personal privacy and security. So, it is urgent to develop effective facial privacy protection methods against the invasive use by unauthorized systems.\par

One of the most widely used ways to protect facial privacy is targeted de-identification (impersonation), which conceals the original identity by generating a protected image to impersonate a real or synthesized target identity. An ideal protected image must (1) perceive high visual quality, ensuring that humans can still recognize the same identity as in the original image, and (2) provide effective privacy protection by successfully concealing the original identity from unauthorized FR systems \cite{Clip2protect_2023_CVPR}.\par

To this end, TIP-IM \cite{TIP-IM_2021_CVPR} learns noise-constrained adversarial perturbations to conceal the original identity. However, the learned noise mask is perceptible, adversely affecting the generated image quality, as shown in Fig. \ref{fig_1}(a). Instead, makeup-based methods utilize adversarial makeup to produce semantically meaningful distortions for privacy protection. For example, AMT-GAN \cite{AMT-GAN_2022_CVPR} and DiffAM \cite{DiffAM_2024_CVPR} transfers the fine-grained makeup style from a reference image to the original one while impersonating a target identity. However, they need an extra reference image to extract the makeup information and have to retrain the whole model for each new target identity, which makes them less practical and time-consuming (see Fig. \ref{fig_1}(b) and (c)). Differently, Clip2Protect \cite{Clip2protect_2023_CVPR} uses pre-defined textual makeup prompts to add makeup to the original image. Still, it struggles to achieve precise control over fine-grained aspects like makeup placement and color, often resulting in discrepancies between the generated makeup and the intended prompt (see Fig. \ref{fig_1}(d)). For makeup-based methods, improving the protection performances often requires applying more makeup across the face or intensifying it in a specific area, adversely affecting the overall image quality.\par

Driven by the powerful image generation of Diffusion Autoencoder (Diff-AE) \cite{Diff_autoencoders_2022_CVPR}, Liu et al, as the pioneers, propose DiffProtect \cite{Diffprotect_2023_arXiv} which employs the frozen Diff-AE model to generate protected face images. In detail, DiffProtect encodes the input image as a semantic code via a semantic encoder and a stochastic code through a conditional DDIM encoding process, then adversarially optimizes an adversarial semantic code to create semantically meaningful perturbations. However, modifying the semantic code makes the protected face's shape and structure resemble the target face, resulting in perceivable distortions, as shown in Fig. \ref{fig_1}(e). More seriously, DiffProtect suffers from \textbf{the diffusion purification effect} leading to an unsatisfying protection performance. Diffusion purification naturally occurs in the reverse diffusion process \cite{carlini2023certified}, where adversarial modifications are interpreted as high-frequency noise and are gradually removed throughout the denoising steps \cite{Diff_purification_2022_ICML}.\par

To address the mentioned issues, we propose learning unconditional embeddings during the reverse diffusion process and then directly optimizing an adversarial latent code within a latent diffusion model (LDM) \cite{Stable_diff_2022_CVPR}. This approach eliminates the need to modify the semantic latent code to generate a protected face image that preserves the overall structures. On the one hand, these learnable unconditional embeddings provide extra learning capacity to the model, allowing it to retain fine textures and structural details. On the other hand, they can effectively preserve identity-related alterations by weakening the model from overly purifying the adversarial modifications. This allows adversarial features to persist in the generated image, thereby simultaneously improving the generated image's visual quality and protection capability. Although learning unconditional embeddings aids in generating images with good visual quality, unrestricted modification of the latent code reduces the structural consistency between the original and generated images. Consistency in the self-attention maps before and after latent code learning is proposed to make the generated face image have the same structure as the original face. The overall pipeline for our method is illustrated in Fig. \ref{fig_2}. In summary, key contributions are summarized below:
\begin{itemize}
\item{We propose a novel framework for facial privacy protection that adversarially modifies the latent code in LDM and introduces learned unconditional embeddings as null-text guidance to guide the generation of protected images. Benefiting from the null-text guidance, our method can weaken the diffusion purification and retain more fine textures and structure details to enhance both the generated image's protection capability and visual quality.}
\item{We propose to leverage self-attention maps to preserve the structural integrity between the original and generated images, thereby maintaining the visual quality of the protected images.}
\item{Extensive experiments on the CelebA-HQ \cite{CelebA-HQ_2017_arXiv} and LADN \cite{LADN_2019_ICCV} datasets demonstrate the efficacy of our method in protecting facial privacy by impersonating real and synthesized face images, with notable improvements in protection success rates (PSR) and Fréchet inception distance (FID) \cite{FID_2017_ANIPS}.}
\end{itemize}\par
\section{Related Work}
\label{sec:Related}
Facial privacy protection methods can be categorized as white-box \cite{white-box_attack_2021_Multimedia}, where the parameters and architectures of the target model are accessible, or black-box, where only limited information about the target model is available. As our work belongs to transfer-based black-box methods \cite{MI-FGSM_2018_CVPR}, where protected face images are generated using surrogate FR models in a white-box setting to deceive an unauthorized FR model, we review the black-box methods. They can be categorized into noise-based, patch-based, makeup-based, and diffusion-based methods. In what follows, we elaborately review each category.\par

\textbf{Noise-based Method.} Noise-based methods generate unexplainable adversarial noises on facial images to conceal their identities. Several approaches have been proposed, such as projected gradient descent (PGD) \cite{PGD_2017_arXiv}, momentum iterative fast gradient sign method (MI-FGSM) \cite{MI-FGSM_2018_CVPR}, translation-invariant diverse inputs method (TI-DIM) \cite{TI-DIM_2019_CVPR}, and targeted identity-protection iterative method (TIP-IM) \cite{TIP-IM_2021_CVPR}. Although these methods effectively protect privacy, they often produce output images with noticeable noises, affecting the overall user experience.

\textbf{Patch-based Method.} Patch-based de-identification has been developed to create wearable accessories, such as colorful glasses in Adv-Glasses \cite{Adv-Glasses_2019_ACM-TOPS} and hats in Adv-Hat \cite{Adv-Hat_2021_ICPR}, to protect facial privacy. Xiao \etal \cite{Patch_2021_CVPR} extend previous transfer-based approaches, create adversarial patches, and regularize them on a low-dimensional data manifold represented by generative models to improve their transferability. However, these adversarial patches often suffer from poor protection performance due to their confined editing region, reducing the image's natural appearance.

\textbf{Makeup-based Method.} More recent methods protect facial privacy by applying makeup. For instance, Adv-Makeup \cite{Adv-makeup_2021_IJCAI} and AMT-GAN \cite{AMT-GAN_2022_CVPR} focus on makeup transfer. A subset of style transfer, makeup transfer is an image-to-image translation task \cite{image-translation_2017_CVPR} that blends the content of a source image with the makeup style of a reference image. Another example, CLIP2Protect \cite{Clip2protect_2023_CVPR}, utilizes a CLIP-guided \cite{CLIP_2021_ICML} model to apply makeup by modifying latent codes in StyleGAN \cite{Auto-StyleGAN_2021_ACM-Trans}. All these methods add makeup using GANs, which tend to introduce unexpected makeup artifacts and struggle to maintain non-makeup elements, such as the background. This often results in face images that appear visually unnatural to human observers. To improve the quality of the protected face image, DiffAM \cite{DiffAM_2024_CVPR} employs two diffusion models, one for makeup removal guided by text and one for adversarial makeup transfer guided by a reference makeup image, respectively. Nevertheless, one key drawback remains. The model must be retrained whenever the target identity changes in the targeted de-identification.

\textbf{Diffusion-based Method.} As cutting-edge probabilistic generative models, diffusion models \cite{DDPM_2020_ANIPS, DDIM_2021_ICLR, Diff-vs-GAN_2021_ANIPS, Improved-DDPM_2021_ICML} excel in producing highly realistic and high-resolution images. Research has recently shown that their strong generation capabilities can be leveraged to generate adversarial examples. For instance, DiffProtect \cite{Diffprotect_2023_arXiv} demonstrated that modifying the semantic latent codes in Diff-AE \cite{Diff_autoencoders_2022_CVPR} enables the generation of protected images. However, modifying these codes can significantly alter the face’s structure, compromising structural consistency between the original and generated images. DiffProtect constrains the modification strength of the semantic code to mitigate the structure distortion but sacrifices the protection capability. Moreover, pre-trained diffusion models have proven to be robust purification tools \cite{Diff-robustness_2021_ANIPS, Diff_purification_2022_ICML}, effectively removing adversarial noise from input images. DiffProtect ignores the adverse influences of the purification effect in the reverse diffusion process, further limiting its protection capability. Instead of modifying a semantic latent code in Diff-AE, our method directly and unrestrictedly modifies the latent codes in LDM \cite{Stable_diff_2022_CVPR} and uses the learned unconditional embeddings as null-text guidance to guide the generation of protected images and weaken the diffusion purification effect. Besides, we preserve the structural consistency by aligning similarity between self-attention maps rather than constraining the modification strength on the adversarial code.

\textbf{Diffusion Purification.} Diffusion‐based models provide a promising approach to adversarial purification \cite{Diff_purification_2022_ICML, purify++_2023_arXive, robust_purification_2023_ICCV} by considering imperceptible adversarial perturbations as noise. The purification process involves a forward diffusion stage that progressively adds noise to an adversarial example over $t$ steps, and a reverse denoising stage that removes this noise to recover the clean data. Theoretically, as more noise is added, the distributions of the noisy adversarial and true examples become increasingly similar \cite{Diff_purification_2022_ICML}, making the denoised outputs likely to converge toward the clean data. It has been shown that the purified data share a similar distribution with the clean data \cite{purify++_2023_arXive}, effectively erasing the adversarial attack.
\section{Method}
\label{sec:method}

\begin{figure*}[t]
    \centering
    \includegraphics{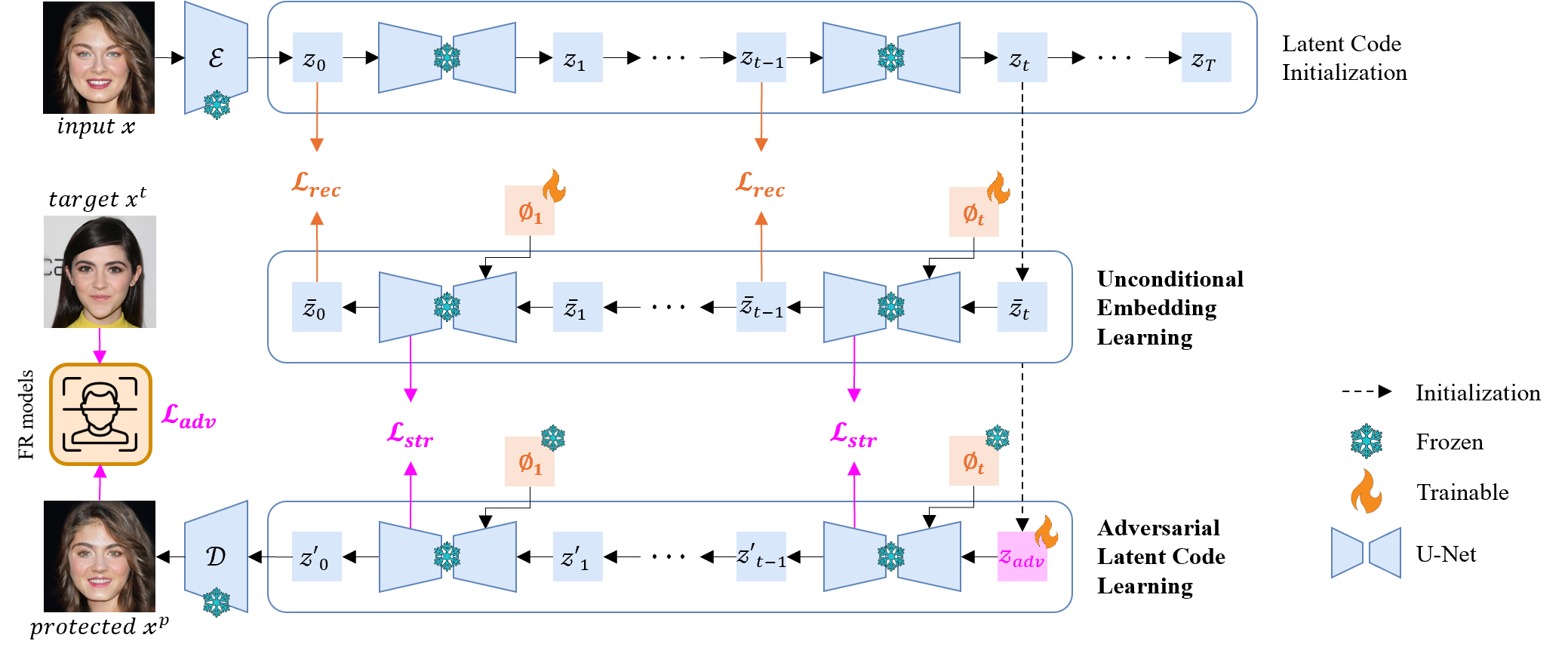}%
    \caption{The overview of the proposed framework for facial privacy protection. Our novel approach leverages Stable Diffusion \cite{Stable_diff_2022_CVPR} to adversarially modify the latent code $z_{adv}$, enabling subtle and controlled alterations to identity-specific features, ensuring effective facial privacy protection while maintaining high visual quality. Unconditional embeddings are proposed as null-text guidance to weaken diffusion purification and enhance protection performance. Self-attention guidance is employed to preserve the structural integrity of the image, ensuring the generated faces remain visually consistent with the original while maintaining high protection efficacy.}
    \label{fig_2}
\end{figure*}

\subsection{Problem Definition \& Framework Overview}
Following previous work \cite{Clip2protect_2023_CVPR, DiffAM_2024_CVPR, Diffprotect_2023_arXiv}, we outline the problem of targeted de-identification in this section. Suppose the original face image $x$ is given, the goal is to generate a protected face image $x^{p}$ that impersonates the target face image $x^{t}$ while maintaining the natural appearance and overall facial identity of the original image $x$ as human observers perceive. Formally, facial privacy protection can be defined as the following optimization problem:
\begin{equation}
    \label{eq1}
    \min _{x^{p}} \mathcal{L}_{a d v}=\mathcal{D}\left(\mathcal{F}\left(x^{p}\right), \mathcal{F}\left(x^{t}\right)\right)\\
    \text { s.t. } \mathcal{P}\left({x}^{p}, {x}\right) \leq \tau\;,
\end{equation}
where $\mathcal{F}$ denotes the feature extractor used in the FR model, $\mathcal{D}$ is a distance function (cosine distance in our work), $\mathcal{P}$ is a measure of the perceptual difference between the original and the protected images, and $\tau$ is a threshold.\par

As illustrated in Fig. \ref{fig_2}, our proposed method leverages the pre-trained diffusion model and a two-stage learning strategy to generate the protected face image that effectively conceals the identity of the original image so that it could not be identified by unauthorized FR systems, and maintains high visual quality for human observers. With the input image $x$, we first feed it to the diffusion model to obtain its noise vector $z_t$. Then we learn per-timestamp unconditional embeddings $\{\varnothing_i\}_{i=1}^{t}$ with two main objectives: (1) to achieve high-quality protected images, and (2) to weaken the purification effect in the reverse diffusion process. In the second stage, we freeze the unconditional embedding learned in the first stage and modify the adversarial latent code $z_{adv}$ to generate the protected face image by minimizing two loss items: the adversarial loss which ensures identity concealment and the structure preservation loss which ensures that the generated image retains key structural features. The details of each part are described as follows.

\subsection{Latent Code Initialization}
Unlike DDPM \cite{DDPM_2020_ANIPS} and DDIM \cite{DDIM_2021_ICLR}, which operate directly in pixel space, the conditional latent diffusion model (LDM) \cite{Stable_diff_2022_CVPR} operates within a compressed latent space, learned through an autoencoder to reduce computational complexity. The autoencoder encodes the input image $x$ into a latent representation $z=\mathcal{E}(x)$ and decodes it back into $\hat{x}=\mathcal{D}(z)$.\par

Since we aim to edit a given real image $x$, it is necessary to identify a noise map $z_{t}$ that reconstructs the input latent code $z_{0}$ satisfying $z_{0}=x$ during the sampling process. Owing to the deterministic nature of DDIM and the fact that it models the reverse diffusion as an ordinary differential equation, it is possible to map a real image $z_0$ back to its corresponding latent representation ($z_{1}, ..., z_{t}$) by reversing the sampling process, a technique known as DDIM Inversion \cite{Null-text_2023_CVPR}:
\begin{multline}
    \label{eq2}
    z_{t+1}=\sqrt{\bar{\alpha}_{t+1} / \bar{\alpha}_{t}} z_{t} + \\\sqrt{\bar{\alpha}_{t+1}}\left(\sqrt{1 / \bar{\alpha}_{t+1}-1}-\sqrt{1 / \bar{\alpha}_{t}-1}\right) \epsilon_{\theta}\left(z_{t}, t, \varnothing\right)\;,
\end{multline}
\noindent where $\bar{\alpha}_{t}$ is the noise scaling factor and $\epsilon_{\theta}\left(z_{t}, t, \varnothing\right)$ is the U-Net model that predicts the noise added at timestamp $t$, given the unconditional embedding $\varnothing$. Since LDM \cite{Stable_diff_2022_CVPR} relies on conditions, while our approach does not depend on textual prompts as conditional input, we use a single unconditional embedding $\varnothing$, initialized with a null-text embedding, during the inversion process. DDIM sampling process is also shown below:
\begin{multline}
    \label{eq3}
    z_{t-1}=\sqrt{\bar{\alpha}_{t-1} / \bar{\alpha}_{t}} z_{t} + \\\sqrt{\bar{\alpha}_{t-1}}\left(\sqrt{1 / \bar{\alpha}_{t-1}-1}-\sqrt{1 / \bar{\alpha}_{t}-1}\right) \epsilon_{\theta}\left(z_{t}, t, \varnothing_t\right)\;.
\end{multline}\par

\subsection{Unconditional Embedding Learning}
We observed that inversion using DDIM can lead to degradation of the regenerated image quality. This process often results in a reconstructed image with increased smoothness, creating a more even texture, particularly noticeable on the skin and hair strands. An intuitive way to invert real images into the model's domain is to fine-tune the model weights for each input image \cite{Auto-StyleGAN_2021_ACM-Trans}. Still, this method is highly inefficient and computationally expensive. Another approach used in various editing tasks is optimizing the textual prompts \cite{Prompt-to-Prompt_2023_ICLR, Instructpix2pix_CVPR_2023}. However, our proposed method is text-free and we observed that fine-grained details tend to diminish without textual guidance during the reverse process.\par

To address this, we propose to learn unconditional embeddings, inspired by the method introduced by Mokady et al \cite{Null-text_2023_CVPR}. As illustrated in Fig. \ref{fig_2}, we apply the DDIM sampling step (Eq. \ref{eq3}) using $\bar{z}_{t}=z_{t}$  with $\varnothing_{t}$ as the model's condition to obtain $\bar{z}_{t-1}$. To enforce that the reconstructed image is as similar to the original one while preserving the capacity for meaningful edits, we enforce $\bar{z}_{t-1}$ approximate to $z_{t-1}$ via:
\begin{equation}
    \label{eq4}
    \min _{\varnothing_{t}} \mathcal{L}_{rec}=\left\|z_{t-1}-\bar{z}_{t-1}\right\|_{2}^{2}.
\end{equation}
We initiate the optimization process at a specific timestamp $t$, where adversarial optimization begins, and optimize a distinct unconditional embedding for each timestamp. This allows us to maximize the similarity to the original image while preserving the capacity for meaningful edits.\par

\begin{figure}[!t]
    \centering
    \resizebox{0.95\columnwidth}{!}
    {\begin{tabular}{cccc}
        \begin{subfigure}[b]{0.2\textwidth}
            \includegraphics[width=\linewidth]{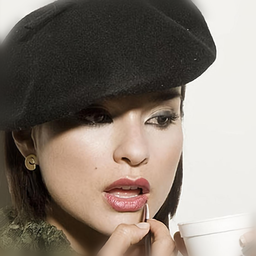}
            \centering
            \vspace{-15pt}
            \caption*{\colorbox{black}{\parbox{\dimexpr\linewidth-2\fboxsep}{\centering\textbf{\fontsize{10pt}{12pt}\selectfont\textcolor[rgb]{1, 0.75, 0}{36.33}}}}}
        \end{subfigure} &
        \begin{subfigure}[b]{0.2\textwidth}
            \includegraphics[width=\linewidth]{images/fig1/015233.png}
            \centering
            \vspace{-15pt}
            \caption*{\colorbox{black}{\parbox{\dimexpr\linewidth-2\fboxsep}{\centering\textbf{\fontsize{10pt}{12pt}\selectfont\textcolor[rgb]{1, 0.75, 0}{38.58}}}}}
        \end{subfigure} &
        \begin{subfigure}[b]{0.2\textwidth}
            \includegraphics[width=\linewidth]{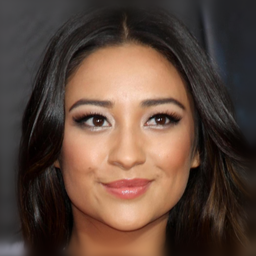}
            \centering
            \vspace{-15pt}
            \caption*{\colorbox{black}{\parbox{\dimexpr\linewidth-2\fboxsep}{\centering\textbf{\fontsize{10pt}{12pt}\selectfont\textcolor[rgb]{1, 0.75, 0}{56.14}}}}}
        \end{subfigure} &
        \begin{subfigure}[b]{0.2\textwidth}
            \includegraphics[width=\linewidth]{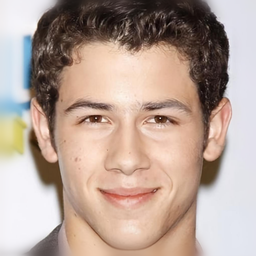}
            \centering
            \vspace{-15pt}
            \caption*{\colorbox{black}{\parbox{\dimexpr\linewidth-2\fboxsep}{\centering\textbf{\fontsize{10pt}{12pt}\selectfont\textcolor[rgb]{1, 0.75, 0}{32.51}}}}}
        \end{subfigure} \\

        \begin{subfigure}[b]{0.2\textwidth}
            \includegraphics[width=\linewidth]{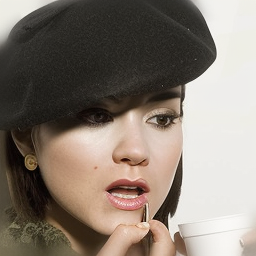}
            \centering
            \vspace{-15pt}
            \caption*{\colorbox{black}{\parbox{\dimexpr\linewidth-2\fboxsep}{\centering\textbf{\fontsize{10pt}{12pt}\selectfont\textcolor[rgb]{1, 0.75, 0}{69.34}}}}}
        \end{subfigure} &
        \begin{subfigure}[b]{0.2\textwidth}
            \includegraphics[width=\linewidth]{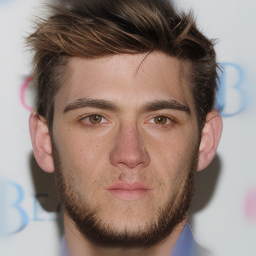}
            \centering
            \vspace{-15pt}
            \caption*{\colorbox{black}{\parbox{\dimexpr\linewidth-2\fboxsep}{\centering\textbf{\fontsize{10pt}{12pt}\selectfont\textcolor[rgb]{1, 0.75, 0}{70.60}}}}}
        \end{subfigure} &
        \begin{subfigure}[b]{0.2\textwidth}
            \includegraphics[width=\linewidth]{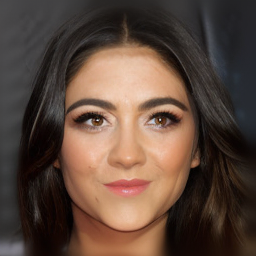}
            \centering
            \vspace{-15pt}
            \caption*{\colorbox{black}{\parbox{\dimexpr\linewidth-2\fboxsep}{\centering\textbf{\fontsize{10pt}{12pt}\selectfont\textcolor[rgb]{1, 0.75, 0}{87.88}}}}}
        \end{subfigure} &
        \begin{subfigure}[b]{0.2\textwidth}
            \includegraphics[width=\linewidth]{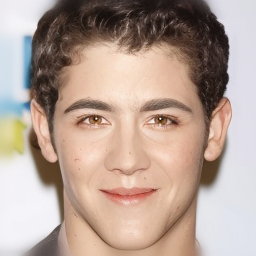}
            \centering
            \vspace{-15pt}
            \caption*{\colorbox{black}{\parbox{\dimexpr\linewidth-2\fboxsep}{\centering\textbf{\fontsize{10pt}{12pt}\selectfont\textcolor[rgb]{1, 0.75, 0}{75.95}}}}}
        \end{subfigure} \\

        \begin{subfigure}[b]{0.2\textwidth}
            \includegraphics[width=\linewidth]{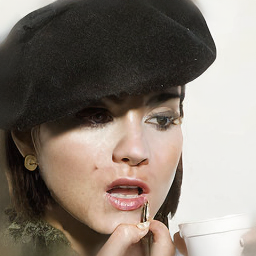}
            \centering
            \vspace{-15pt}
            \caption*{\colorbox{black}{\parbox{\dimexpr\linewidth-2\fboxsep}{\centering\textbf{\fontsize{10pt}{12pt}\selectfont\textcolor[rgb]{1, 0.75, 0}{75.99}}}}}
        \end{subfigure} &
        \begin{subfigure}[b]{0.2\textwidth}
            \includegraphics[width=\linewidth]{images/fig1/015233_prot.png}
            \centering
            \vspace{-15pt}
            \caption*{\colorbox{black}{\parbox{\dimexpr\linewidth-2\fboxsep}{\centering\textbf{\fontsize{10pt}{12pt}\selectfont\textcolor[rgb]{1, 0.75, 0}{73.88}}}}}
        \end{subfigure} &
        \begin{subfigure}[b]{0.2\textwidth}
            \includegraphics[width=\linewidth]{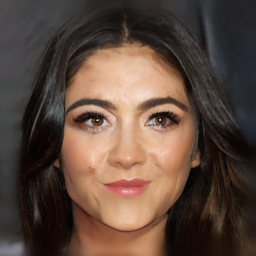}
            \centering
            \vspace{-15pt}
            \caption*{\colorbox{black}{\parbox{\dimexpr\linewidth-2\fboxsep}{\centering\textbf{\fontsize{10pt}{12pt}\selectfont\textcolor[rgb]{1, 0.75, 0}{89.25}}}}}
        \end{subfigure} &
        \begin{subfigure}[b]{0.2\textwidth}
            \includegraphics[width=\linewidth]{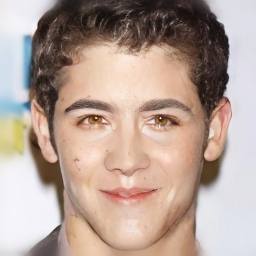}
            \centering
            \vspace{-15pt}
            \caption*{\colorbox{black}{\parbox{\dimexpr\linewidth-2\fboxsep}{\centering\textbf{\fontsize{10pt}{12pt}\selectfont\textcolor[rgb]{1, 0.75, 0}{78.26}}}}}
        \end{subfigure} \\
    \end{tabular}}
    \caption{The improvements gained by incorporating learned unconditional embeddings during adversarial image generation. The first row presents original images. The second row shows protected images generated without unconditional embeddings as the null-text guidance. The third row displays protected images generated with our learned unconditional embeddings as the null-text guidance. The protection capability of the protected images is enhanced with the null-text guidance.}
    \label{fig_3}
\end{figure}

\subsection{Adversarial Latent Code Learning}
In a black-box setting, where the malicious FR model is unknown, the optimization problem cannot be solved directly. Therefore, like previous methods, we perform adversarial optimization on $K$ white-box surrogate models to closely approximate the unknown FR model's decision boundaries. Starting with $z_{adv}=z_{t}$, multiple DDIM reverse steps are applied to obtain $z_{0}^{\prime}$, which is then decoded as $x^{p}=\mathcal{D}\left(z_{0}^{\prime}\right)$, as illustrated in Fig. \ref{fig_2}. The loss function is defined as follows:
\begin{equation}
    \label{eq5}
    \min _{z_{adv}} \mathcal{L}_{\text {adv}}=\frac{1}{K} \sum_{k=1}^{K}\left[1-\cos \left(\mathcal{F}_{k}\left(x^{p}\right), \mathcal{F}_{k}\left(x^{t}\right)\right)\right].
\end{equation}
As stated earlier, $\mathcal{F}_k$ refers to the feature extractor used in the $k$-th FR model.\par

While adversarial modifications in the latent code $z_{adv}$ are intended to generate a transferable protected face image, experimental results reveal that the diffusion model’s reverse process naturally removes these small, high-frequency identity-related modifications \cite{Frequencies-adv-trans_2024_arXiv} due to its strong purification ability \cite{Diff_purification_2022_ICML}. In other words, these perturbations result in a modified $z_{adv}$ that no longer precisely corresponds to the clean image. However, during the reverse process, the denoising model—trained exclusively on clean images—projects the perturbed noise back toward the natural data manifold, thereby mitigating the impact of the adversarial modification. Incorporating learned unconditional embeddings as null-text guidance in the diffusion reverse process, we find that the diffusion purification effect is weakened and identity-specific modifications are effectively preserved. Correspondingly, the privacy protection capability of the protected images is enhanced. The PSR and the visual aspects in Fig. \ref{fig_3} further show the effectiveness of null-text guidance, which encourages the model to maintain certain input details—including the adversarial noise—rather than fully denoising the image.

\subsection{Structure Preservation via Self-attention}
\begin{figure}[!t]
    \centering
    \resizebox{\columnwidth}{!}
    {\begin{tabular}{cccc}
        \begin{subfigure}[b]{0.2\textwidth}
            \centering
            \caption*{\textbf{\fontsize{14pt}{16pt}\selectfont Original Image}}
            \vspace{2pt}
            \includegraphics[width=\linewidth]{images/fig1/000737.png}
        \end{subfigure} &
        \begin{subfigure}[b]{0.2\textwidth}
            \centering
            \caption*{\textbf{\fontsize{14pt}{16pt}\selectfont Top-0}}
            \vspace{2pt}
            \includegraphics[width=\linewidth]{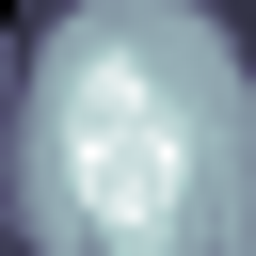}
        \end{subfigure}
        \begin{subfigure}[b]{0.2\textwidth}
            \centering
            \caption*{\textbf{\fontsize{14pt}{16pt}\selectfont Top-1}}
            \vspace{2pt}
            \includegraphics[width=\linewidth]{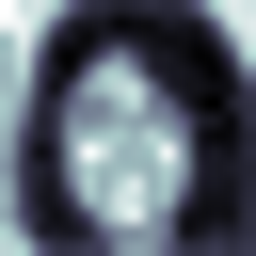}
        \end{subfigure}
        &
        \raisebox{7.2\height}{\textbf{\fontsize{14pt}{16pt}\selectfont $\cdots$}}
        &
        \begin{subfigure}[b]{0.2\textwidth}
            \centering
            \caption*{\textbf{\fontsize{14pt}{16pt}\selectfont Top-5}}
            \vspace{2pt}
            \includegraphics[width=\linewidth]{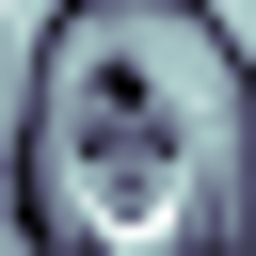}
        \end{subfigure}
        \begin{subfigure}[b]{0.2\textwidth}
            \centering
            \caption*{\textbf{\fontsize{14pt}{16pt}\selectfont Top-6}}
            \vspace{2pt}
            \includegraphics[width=\linewidth]{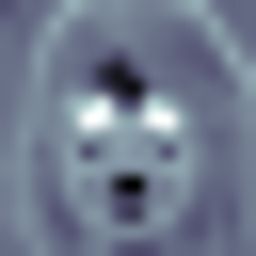}
        \end{subfigure} \\
        \begin{subfigure}[b]{0.2\textwidth}
            \centering
            \caption*{\textbf{\fontsize{14pt}{16pt}\selectfont Protected Image}}
            \vspace{2pt}
            \includegraphics[width=\linewidth]{images/fig1/000737_prot.png}
        \end{subfigure} &
        \begin{subfigure}[b]{0.2\textwidth}
            \centering
            \includegraphics[width=\linewidth]{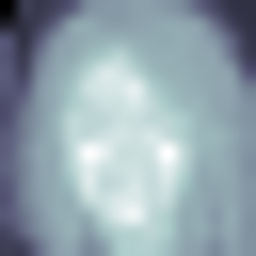}
        \end{subfigure}
        \begin{subfigure}[b]{0.2\textwidth}
            \centering
            \includegraphics[width=\linewidth]{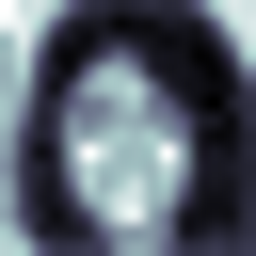}
        \end{subfigure}
        &
        \raisebox{7.2\height}{\textbf{\fontsize{14pt}{16pt}\selectfont $\cdots$}}
        &
        \begin{subfigure}[b]{0.2\textwidth}
            \centering
            \includegraphics[width=\linewidth]{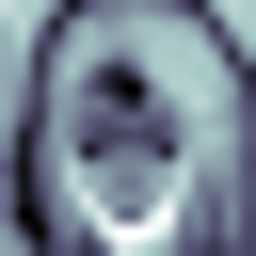}
        \end{subfigure}
        \begin{subfigure}[b]{0.2\textwidth}
            \centering
            \includegraphics[width=\linewidth]{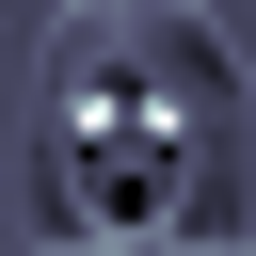}
        \end{subfigure} \\
    \end{tabular}}
    \caption{The self-attention map \cite{Transformers_2017_ANIPS} visualization shows the top seven components extracted through singular value decomposition \cite{SVD_2003_Springer}. The above shows the self-attention maps corresponding to the original image, while the maps for the protected image are shown below. Despite the protection, the structural details of the original image are well preserved in the protected images.}
    \label{fig_4}
\end{figure}

While learned unconditional embeddings can partially retain the quality of the generated image, unrestricted optimization of the latent code $z_{adv}$ (Eq. \ref{eq5}) may alter the structure of the generated image, causing it to resemble the target image more closely than the original. A potential solution to this issue is to constrain the latent code to meet the $L_{\infty}$ norm bound $\left\|z_{adv}-z_{t}\right\|_{\infty} \leq \varepsilon$ \cite{MI-FGSM_2018_CVPR}. However, this contradicts our unrestricted approach and leads to weaker protection. Research by \cite{Zero-shot-translation_2023_ACM-SIGGRAPH} has revealed that differences in attention maps between edited and real images lead to structural changes in the edited image. \cite{Cross-Self-Attention_2024_CVPR} further shows that refining the cross-attention maps of an adversarial image using the source image during the generation process can lead to unsuccessful image editing. Moreover, modifying cross-attention maps typically relies on a conditional prompt, as discussed in Section 3.3, our approach does not require any text input. In contrast, self-attention maps \cite{Transformers_2017_ANIPS} are essential for preserving the geometric and shape details of the source image (see Fig. \ref{fig_4}). Therefore, we propose using self-attention guidance to retain the image structure during latent code learning.\par

We adopt a two-step process \cite{DiffAttack_trans_2024}, as shown in Fig. \ref{fig_2}. First, we apply the DDIM sampling steps using the latent code $\bar{z}_{t}$ before applying any perturbations and calculate self-attention maps $S(\bar{z}_{t})$ for each timestamp $t$. These maps capture the original image's structure, which we aim to preserve. Then, we calculate self-attention maps $S({z}_{adv})$ after adding the perturbation again for each timestamp. We align $S({z}_{adv})$ and $S(\bar{z}_{t})$ by minimizing their difference:
\begin{equation}
    \label{eq6}
    \min _{z_{adv}} \mathcal{L}_{\text {str }}=\left\|S\left(z_{adv}\right) - S\left(\bar{z}_t\right)\right\|_{2}^{2}.
\end{equation}
Overall, the final loss function for our proposed method is defined as follows:
\begin{equation}
    \label{eq7}
    \min _{z_{adv}} \mathcal{L}=\lambda_{\text {adv }} \mathcal{L}_{\text {adv }}+\mathcal{L}_{\text {str }},
\end{equation}
where $\lambda_{a d v}$ represents the hyper-parameter.

\section{Experiments}
\label{sec:experiment}
\subsection{Experimental Setting}
\textbf{Datasets.} Following AMT-GAN \cite{AMT-GAN_2022_CVPR}, we employ the CelebA-HQ \cite{CelebA-HQ_2017_arXiv} and LADN \cite{LADN_2019_ICCV} datasets in our experiments. CelebA-HQ, a high-resolution variant of the CelebA dataset, contains 30,000 images. Following \cite{AMT-GAN_2022_CVPR}, a subset of 1000 images is selected, representing distinct identities. The LAND dataset includes 333 non-makeup and 302 makeup images, and based on \cite{AMT-GAN_2022_CVPR}, we select 332 non-makeup images for testing. In both datasets, the selected images are divided into four groups, each corresponding to a specific target identity. All images within a group aim to impersonate the same target identity, following the four target identities provided by \cite{AMT-GAN_2022_CVPR}.\par
\textbf{Target models.} We carry out extensive experiments with four widely-used black-box FR models—IRSE50 \cite{IRSE50_2018_CVPR}, IR152 \cite{IR152_2016_CVPR}, Facenet \cite{Facenet_2015_CVPR}, and Mobileface \cite{Mobilefacenets_2018}—to evaluate the protection performance. We select three models during fine-tuning, while the remaining one is reserved for black-box testing. Following the experimental protocol outlined in \cite{Clip2protect_2023_CVPR}, we preprocess the face images by aligning and cropping them using MTCNN \cite{MTCNN_2016_IEEE-SP} before inputting them into the FR models for evaluation.\par
\textbf{Benchmarks.} We compare our method with four noise-based, four makeup-based, and one diffusion-based facial privacy protection techniques. Noise-based methods include PGD \cite{PGD_2017_arXiv}, MI-FGSM \cite{MI-FGSM_2018_CVPR}, TI-DIM \cite{TI-DIM_2019_CVPR}, and TIP-IM \cite{TIP-IM_2021_CVPR}. Makeup-based methods are Adv-Makeup \cite{Adv-makeup_2021_IJCAI}, AMT-GAN \cite{AMT-GAN_2022_CVPR}, CLIP2Protect \cite{Clip2protect_2023_CVPR}, and DiffAM \cite{DiffAM_2024_CVPR}, and the diffusion-based method is DiffProtect \cite{Diffprotect_2023_arXiv}. DiffAM and DiffProtect are state-of-the-art (SOTA) in makeup-based and diffusion-based de-identification.\par
\textbf{Evaluation metrics.} In line with CLIP2Protect \cite{Clip2protect_2023_CVPR}, we use the PSR to assess our proposed method's effectiveness compared to benchmark approaches. PSR is calculated by setting the false acceptance rate (FAR) to 0.01 \cite{AMT-GAN_2022_CVPR, Clip2protect_2023_CVPR, DiffAM_2024_CVPR}. Additionally, we report the FID \cite{FID_2017_ANIPS}, peak signal-to-noise ratio (PSNR in dB), and structural similarity index (SSIM) \cite{SSIM_2004_IEEE-IP} to evaluate the visual quality of the protected images.\par
\textbf{Implemention Details.} Our implementation is built on \cite{Null-text_2023_CVPR}. We apply 20 DDIM Inversion steps ($T$ = 20) to the initial clean image, with the reverse process beginning at the 3rd timestamp ($t$ = 3). We utilize the AdamW \cite{AdamW_2019_ICLR} optimizer with a learning rate of 0.1 and 20 iterations for unconditional embedding learning. Similarly, we use AdamW with a learning rate of 0.01 and 35 iterations for adversarial latent learning, setting $\lambda_{adv}$ to 0.003.

\begin{table*}[!t]
\centering
\resizebox{\textwidth}{!}
{\begin{tabular}{cl|c|cccc|cccc|c}
\cline{3-12}
\multicolumn{2}{c}{\multirow{3}{*}{\textbf{}}} & \multirow{2}{*}{Method} & \multicolumn{4}{c|}{\textbf{CelebA-HQ}} & \multicolumn{4}{c|}{\textbf{LADN-dataset}} & \multirow{2}{*}{Average} \\ \cline{4-11}
\multicolumn{2}{c}{} & & IRSE50 & IR152 & Facenet & Mobileface & IRSE50 & IR152 & Facenet & Mobileface & \\ 
\cline{3-12} 
\multicolumn{2}{c}{} & clean & 7.29 & 3.80 & 1.08 & 12.68 & 2.71 & 3.61 & 0.60 & 5.11 & 4.61 \\ 
\hline
\multicolumn{2}{c|}{\multirow{4}{*}{Noise-based}} & PGD (2017) & 36.87 & 20.68 & 1.85 & 43.99 & 40.09 & 19.59 & 3.82 & 41.09 & 25.60 \\
\multicolumn{2}{c|}{} & MI-FGSM (2018) & 45.79 & 25.03 & 2.58 & 45.85 & 48.90 & 25.57 & 6.31 & 45.01 & 30.63 \\
\multicolumn{2}{c|}{} & TI-DIM (2019) & 63.63 & 36.17 & 15.30 & 57.12 & 56.36 & 34.18 & 22.11 & 48.30 & 41.64 \\
\multicolumn{2}{c|}{} & TIP-IM (2021) & 54.40 & 37.23  & 40.74 & 48.72 & 65.89 & 43.57 & \underline{63.50} & 46.48 & 50.06 \\ 
\hline
\multicolumn{2}{c|}{\multirow{4}{*}{Makeup-based}} & Adv-Makeup (2021) & 21.95 & 9.48 & 1.37 & 22.00 & 29.64 & 10.03 & 0.97 & 22.38 & 14.72 \\
\multicolumn{2}{c|}{} & AMT-GAN (2022) & 76.96 & 35.13 & 16.62 & 50.71 & 89.64 & 49.12 & 32.13 & 72.43 & 52.84 \\
\multicolumn{2}{c|}{} & CLIP2Protect (2023) & 81.10 & 48.42 & 41.72 & 75.26 & 91.57 & 53.31 & 47.91 & 79.94 & 64.90 \\
\multicolumn{2}{c|}{} & DiffAM (2024) & \textbf{92.00} & \underline{63.13} & \textbf{64.67} & \underline{83.35} & \underline{95.66} & \underline{66.75} & \textbf{65.44} & \underline{92.04} & \underline{77.88} \\ 
\hline
\multicolumn{2}{c|}{\multirow{2}{*}{Diffusion-based}} & DiffProtect (2023) & 67.75 & 60.14 & 35.19 & 64.33 & 54.51 & 44.27 & 31.33 & 50.90 & 51.05 \\
\multicolumn{2}{c|}{} & \textbf{Ours} & \underline{88.87} & \textbf{67.25} & \underline{59.53} & \textbf{91.57} & \textbf{95.78} & \textbf{70.18} & 62.05 & \textbf{98.17} & \textbf{79.17} \\ 
\hline
\end{tabular}}
\caption{Protection success rate (PSR) under the black-box setting. The best and second-best results are marked in bold and underlined.}
\label{tab_1}
\end{table*}

\subsection{Comparison Study}
In this section, we assess the experimental results of our proposed approach and benchmarks in terms of protection performance under black-box settings on four pre-trained FR models, and the generated image quality.\par
\textbf{Assessment on personal facial privacy protection.} The results of black-box settings against four FR models on the CelebA-HQ and LADN datasets are presented in Tab. \ref{tab_1}. The other three models are surrogate models to deceive the target model. We evaluate performance across four target identities and report the average results. It is important to note that the test images of the target faces are different images of the same individuals used during training. Our approach shows about 30\% and 1.5\% improvement in average PSR compared to the SOTA noise-based (TIP-IM) and makeup-based (DiffAM), respectively. Unlike DiffAM, which involves two distinct stages (i.e., makeup removal and makeup transfer), our method employs a more streamlined approach, resulting in better transferability.\par
\textbf{Assessment on image quality.} We perform quantitative and qualitative comparisons for the quality of the generated images. Tab. \ref{tab_2} shows the quantitative evaluation. Among the methods, Adv-Makeup achieves the highest performance across all quality evaluation metrics, focusing solely on applying makeup to the eye region. However, this limited scope leads to a significantly lower PSR. While our method shows lower SSIM performance than DiffAM and DiffProtect, it excels in FID, indicating that the protected images generated are more natural. Additionally, our approach outperforms others in PSNR, suggesting that the controlled modifications made in the latent space have less impact on the images at the pixel level.\par

\begin{table}[h]
\centering
\resizebox{\columnwidth}{!}
{\begin{tabular}{c|c|ccc}
\hline
Method & \textbf{PSR$(\uparrow)$} & \textbf{FID$(\downarrow)$} & \textbf{PSNR$(\uparrow)$} & \textbf{SSIM$(\uparrow)$} \\ \hline
TIP-IM        & 50.06        & 38.7357      & 33.2089       & 0.9214        \\ \hline
Adv-makeup    & 14.72        & 4.2282       & 34.5152       & 0.9850        \\
AMT-GAN       & 52.84        & 34.4405      & 19.5045       & 0.7873        \\
CLIP2Protect  & 64.90        & 37.1172      & 19.3537       & 0.6025        \\
DiffAM        & 77.88        & 26.1015      & 20.5260       & 0.8861        \\ \hline
DiffProtect   & 51.05        & 28.2912      & 24.2070       & 0.8785        \\
\textbf{Ours} & 79.17        & 15.3212      & 27.7223       & 0.8393        \\ \hline
\end{tabular}}
\caption{Quantitative assessments of image quality. These are the average results for both CelebA-HQ and LADN datasets.}
\label{tab_2}
\end{table}

\begin{figure}[!t]
    \centering
    \resizebox{0.85\columnwidth}{!}
    {\begin{tabular}{ccccc}
        \begin{subfigure}[t]{0.2\textwidth}
            \includegraphics[width=\linewidth]{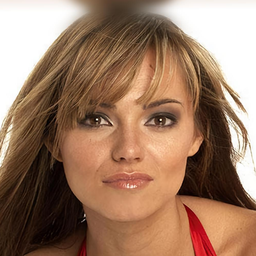}
            \centering
            \vspace{-15pt}
            \caption*{\colorbox{black}{\parbox{\dimexpr\linewidth-2\fboxsep}{\centering\textbf{\fontsize{12pt}{14pt}\selectfont\textcolor[rgb]{1, 0.75, 0}{39.32}}}}}
            \vspace{2pt}
            \fontsize{13pt}{15pt}\selectfont\textcolor{black}{Original Image}
        \end{subfigure}
        \begin{subfigure}[t]{0.2\textwidth}
            \includegraphics[width=\linewidth]{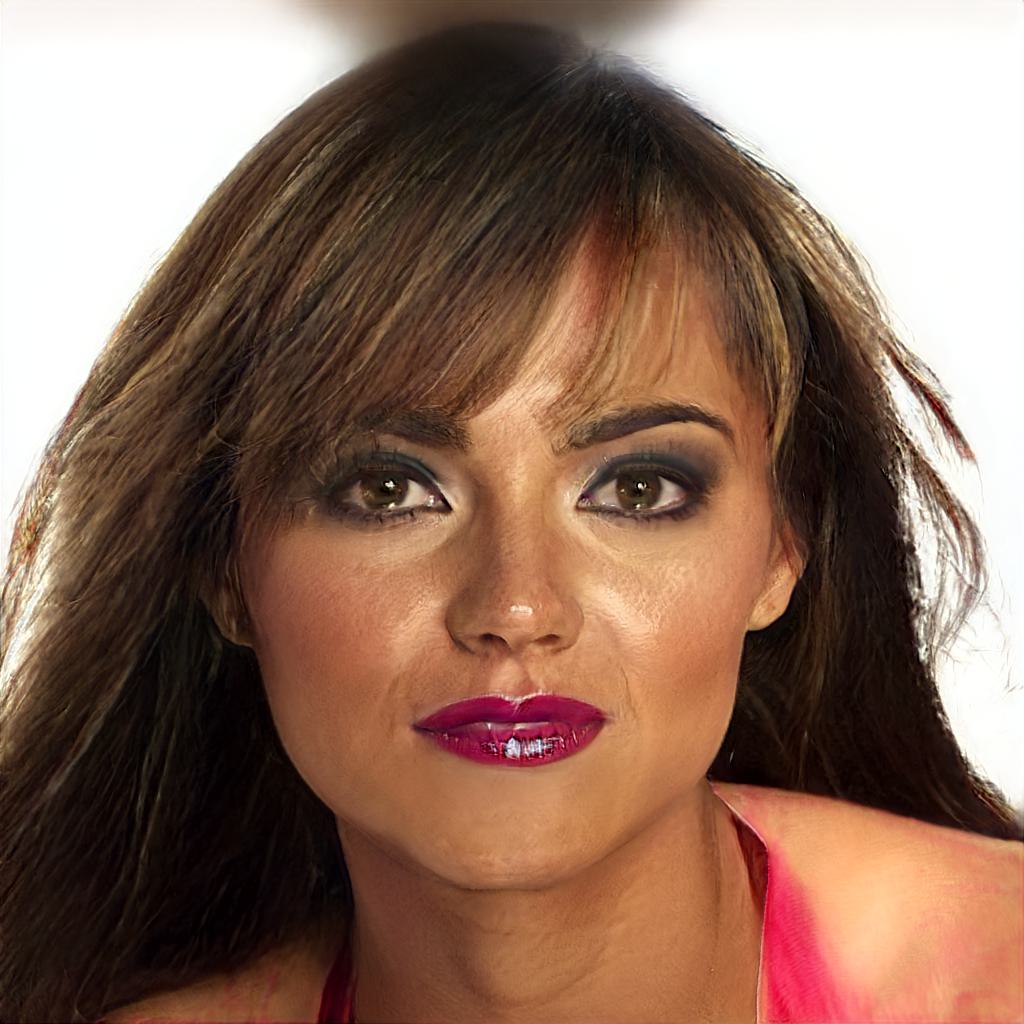}
            \centering
            \vspace{-15pt}
            \caption*{\colorbox{black}{\parbox{\dimexpr\linewidth-2\fboxsep}{\centering\textbf{\fontsize{12pt}{14pt}\selectfont\textcolor[rgb]{1, 0.75, 0}{66.80}}}}}
            \vspace{2pt}
            \includegraphics[width=\linewidth]{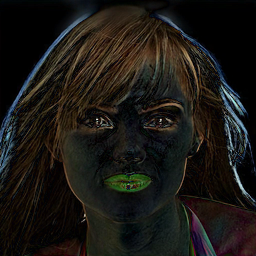}
            \centering
            \vspace{2pt}
            \fontsize{13pt}{15pt}\selectfont\textcolor{black}{CLIP2Protect \cite{Clip2protect_2023_CVPR}}
        \end{subfigure}
        \begin{subfigure}[t]{0.2\textwidth}
            \includegraphics[width=\linewidth]{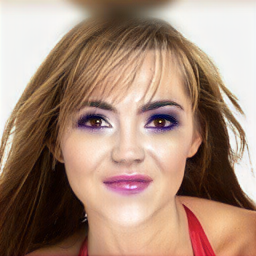}
            \centering
            \vspace{-15pt}
            \caption*{\colorbox{black}{\parbox{\dimexpr\linewidth-2\fboxsep}{\centering\textbf{\fontsize{12pt}{14pt}\selectfont\textcolor[rgb]{1, 0.75, 0}{73.82}}}}}
            \vspace{2pt}
            \includegraphics[width=\linewidth]{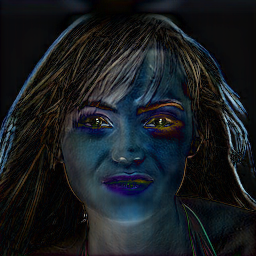}
            \centering
            \vspace{2pt}
            \fontsize{13pt}{15pt}\selectfont\textcolor{black}{DiffAM \cite{DiffAM_2024_CVPR}}
        \end{subfigure}
        \begin{subfigure}[t]{0.2\textwidth}
            \includegraphics[width=\linewidth]{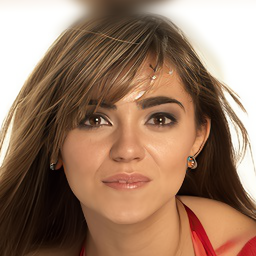}
            \centering
            \vspace{-15pt}
            \caption*{\colorbox{black}{\parbox{\dimexpr\linewidth-2\fboxsep}{\centering\textbf{\fontsize{12pt}{14pt}\selectfont\textcolor[rgb]{1, 0.75, 0}{64.40}}}}}
            \vspace{2pt}
            \includegraphics[width=\linewidth]{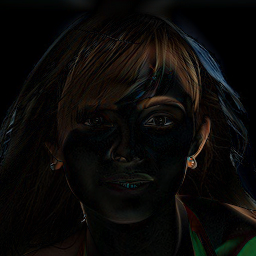}
            \centering
            \vspace{2pt}
            \fontsize{13pt}{15pt}\selectfont\textcolor{black}{DiffProtect \cite{Diffprotect_2023_arXiv}}
        \end{subfigure}
        \begin{subfigure}[t]{0.2\textwidth}
            \includegraphics[width=\linewidth]{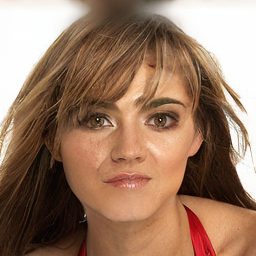}
            \centering
            \vspace{-15pt}
            \caption*{\colorbox{black}{\parbox{\dimexpr\linewidth-2\fboxsep}{\centering\textbf{\fontsize{12pt}{14pt}\selectfont\textcolor[rgb]{1, 0.75, 0}{80.86}}}}}
            \vspace{2pt}
            \includegraphics[width=\linewidth]{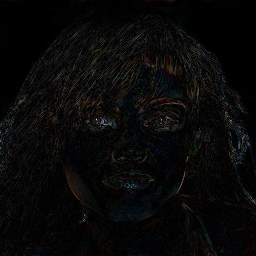}
            \centering
            \vspace{2pt}
            \fontsize{13pt}{15pt}\selectfont\textcolor{black}{Ours}
        \end{subfigure}
    \end{tabular}}
    \caption{Comparison of visual quality between recent facial privacy protection methods. The absolute difference between the generated and original images is shown below each protected face.}
    \label{fig_5}
\end{figure}

The qualitative evaluations of visual quality can be seen in Fig. \ref{fig_5}. The comparison of the generated images reveals notable differences between makeup-based methods (CLIP2Protect and DiffAM), diffusion-based methods (DiffProtect), and the proposed framework. Makeup-based methods apply perturbations across the entire facial region, resulting in more visible alterations in the facial appearance. DiffProtect, while effective, introduces more evenly distributed artifacts that reduce the sharpness and texture of facial features. In contrast, ours achieves superior visual quality by generating more natural-looking faces with minimal, localized perturbations, focusing only on the regions necessary to deceive FR models. This selective approach preserves the overall realism of the faces and avoids introducing noticeable noise patterns, striking an effective balance between adversarial strength and imperceptibility. Consequently, ours outperforms the other methods in maintaining high image fidelity while ensuring protection against FR models.

\subsection{Ablation Studies}

\begin{figure*}[!t]
    \begin{subfigure}[b]{0.4\textwidth}
        \centering
        \begin{tikzpicture}
            \begin{axis}[
                width=0.9\textwidth, 
                height=0.6\textwidth,
                xlabel={timestamp (t)},
                xlabel style={scale=0.7},
                ylabel={PSR},
                ylabel style={scale=0.7},
                xmin=0.75, xmax=7.25,
                ymin=47.5, ymax=102.5,
                xtick={1, 3, 5, 7},
                ytick={50, 60, 70, 80, 90, 100},
                ticklabel style={scale=0.7},
                legend pos=outer north east,
                legend style={nodes={scale=0.65, transform shape}},
                grid=both
            ]
            
            \addplot[
                color=blue,
                mark=*,
                mark size=1.5pt,
                thick
            ]
            coordinates {(1, 93.59) (3, 88.88) (5, 82.36) (7, 74.85)};
            
            \addplot[
                color=orange,
                mark=triangle*,
                mark size=1.5pt,
                thin
            ]
            coordinates {(1, 90.48) (3, 79.86) (5, 67.43) (7, 50.10)};
            
            \addplot[
                color=black,
                mark=+,
                mark size=1.5pt,
                thin
            ]
            coordinates {(1, 95.49) (3, 95.19) (5, 94.39) (7, 94.19)};
            
            \addplot[
                color=red,
                mark=x,
                mark size=1.5pt,
                thin
            ]
            coordinates {(1, 95.29) (3, 94.79) (5, 94.29) (7, 94.09)};
            \legend{w both, w/o null-text, w/o self-attention, w/o both}
            \end{axis}
        \end{tikzpicture}
    \end{subfigure}
    \hspace{2cm}
    \begin{subfigure}[b]{0.4\textwidth}
        \centering
        \begin{tikzpicture}
            \begin{axis}[
                width=0.9\textwidth, 
                height=0.6\textwidth,
                xlabel={timestamp (t)},
                xlabel style={scale=0.7},
                ylabel={FID},
                ylabel style={scale=0.7},
                xmin=0.75, xmax=7.25,
                ymin=7.5, ymax=47.5,
                xtick={1, 3, 5, 7},
                ytick={9, 15, 21, 27, 33, 39, 45},
                ticklabel style={scale=0.7},
                legend pos=outer north east,
                legend style={nodes={scale=0.65, transform shape}},
                grid=both
            ]
            
            \addplot[
                color=blue,
                mark=*,
                mark size=1.5pt,
                thick
            ]
            coordinates {(1, 14.84) (3, 11.37) (5, 10.05) (7, 9.77)};
            
            \addplot[
                color=orange,
                mark=triangle*,
                mark size=1.5pt,
                thin
            ]
            coordinates {(1, 15.08) (3, 11.30) (5, 13.44) (7, 18.60)};
            
            \addplot[
                color=black,
                mark=+,
                mark size=1.5pt,
                thin
            ]
            coordinates {(1, 28.23) (3, 21.94) (5, 21.26) (7, 24.14)};
            
            \addplot[
                color=red,
                mark=x,
                mark size=1.5pt,
                thin
            ]
            coordinates {(1, 30.07) (3, 21.41) (5, 30.66) (7, 43.28)};
            \legend{w both, w/o null-text, w/o self-attention, w/o both}
            \end{axis}
        \end{tikzpicture}
    \end{subfigure}
    \caption{Ablation study to evaluate the contributions of different loss items. The left figure illustrates the protection success rate (PSR) under different conditions, with and without null-text and self-attention guidances, while the right one evaluates the visual quality by Fréchet inception distance (FID). The timestamp $t$ marks the point at which adversarial learning begins.}
    \label{fig_6}
\end{figure*}

\textbf{Null-text guidance.} A comparison of the blue and yellow lines in Fig. \ref{fig_6}(a) and (b) clearly shows how well-learned unconditional embeddings work to make the reverse diffusion process less susceptible to purification and keep the images' visual quality. As we progress to deeper timestamps (e.g., $t = 5$ and $t = 7$), the diffusion model's denoising process prioritizes reconstructing broad structural features over fine-grained details that could aid in impersonation. Moreover, the sampling process's purification effect prevents adversarial modifications from fully remaining, lowering the PSR. With the reverse process beginning at each timestamp, we can see that PSR is much better (particularly at deeper timestamps) when we use these learned embeddings. Looking at the blue and yellow lines in Fig. \ref{fig_6}(b), we can see that these learned embeddings can effectively preserve the quality of generated images by preserving fine-grained details. The trend in the blue line demonstrates that the FID score remains low and relatively stable across timestamps. However, this trend differs in the yellow line in deeper timestamps. It proves the model struggles to maintain fine-grained details without learned embeddings when the starting noise is high.\par
\textbf{Self-attention guidance.} Fig. \ref{fig_6}(a) shows that the lack of self-attention (represented by the black and red lines) leads to higher PSR. However, as seen in Fig. \ref{fig_6}(b), this absence of self-attention also results in a noticeable drop in image quality, indicated by an increase in FID. It can be concluded that without self-attention, the adversarial modifications are less guided, possibly making the image closer to the target identity. At the same time, this lack causes the generated images to deviate more from the original distribution, degrading their visual quality.

\section{Discussion}
\label{sec:discussion}

Generating a protected face by impersonating a real identity raises additional ethical concerns. Although the intent might be to conceal the original identity, using another real identity as the target can misrepresent the individual whose identity is being impersonated. A common approach to address this issue is obfuscation \cite{privacy_pres_2022_CS, obfuscation_2022_FGCS}, which ensures that the protected face does not resemble any other images of the same individual. This can be achieved by generating a protected face where $\mathcal{D}(\mathcal{F}(x^{p}), \mathcal{F}(x))$ is significant. Since the focus is on increasing the distance between the generated and the source face images, the perturbations might not be natural, resulting in distorted images.\par

One way to overcome this limitation is to simultaneously impersonate a target identity via minimizing $\mathcal{D}(\mathcal{F}(x^{p}), \mathcal{F}(x^{t}))$ and obfuscate via maximizing $\mathcal{D}(\mathcal{F}(x^{p}), \mathcal{F}(x))$. As a result, the perturbations are more controlled since impersonation can act as a form of regularization, leading to better visual quality while achieving the obfuscation criteria. Unlike CLIP2Protect \cite{Clip2protect_2023_CVPR}, which impersonates a real target identity in the obfuscation task, we propose impersonating a synthesized target identity to ensure that the protected face does not resemble any real individual. Hence, the protected face maintains anonymity without the risk of misusing a real person’s similarity.\par

\begin{figure}[h]
    \centering
    \resizebox{0.8\columnwidth}{!}
    {\begin{tabular}{cccc}
        \begin{subfigure}[t]{0.2\textwidth}
            \includegraphics[width=\linewidth]{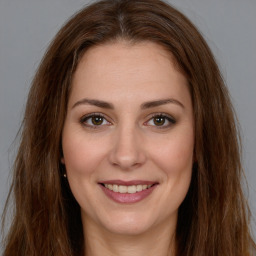}
            \centering
            \vspace{2pt}
            \fontsize{16pt}{18pt}\selectfont{Fake Target}
        \end{subfigure}
        \begin{subfigure}[t]{0.2\textwidth}
            \includegraphics[width=\linewidth]{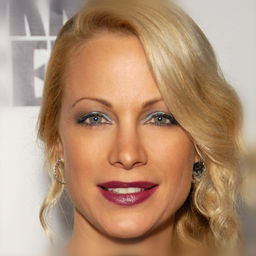}
            \centering
            \vspace{2pt}
            \includegraphics[width=\linewidth]{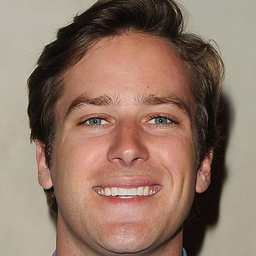}
            \centering
            \vspace{2pt}
            \fontsize{16pt}{18pt}\selectfont{Original Image}
        \end{subfigure}
        \begin{subfigure}[t]{0.2\textwidth}
            \includegraphics[width=\linewidth]{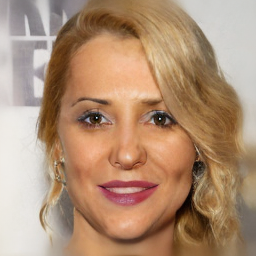}
            \centering
            \vspace{2pt}
            \includegraphics[width=\linewidth]{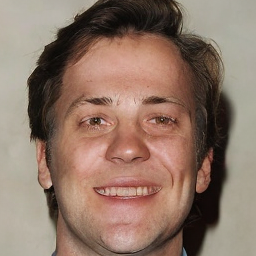}
            \centering
            \vspace{2pt}
            \fontsize{16pt}{18pt}\selectfont{Impersonation}
        \end{subfigure}
        \begin{subfigure}[t]{0.2\textwidth}
            \includegraphics[width=\linewidth]{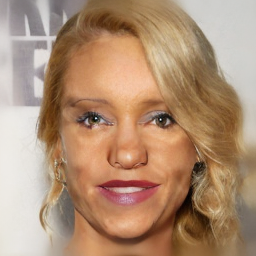}
            \centering
            \vspace{2pt}
            \includegraphics[width=\linewidth]{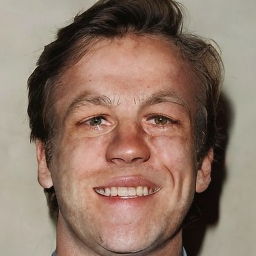}
            \centering
            \vspace{2pt}
            \fontsize{16pt}{18pt}\selectfont{Obfuscation}
        \end{subfigure}
    \end{tabular}}
    \caption{Qualitative assessment of the protected image regarding impersonating a fake target versus merely obfuscation.}
    \label{fig_7}
\end{figure}

We follow CLIP2Protect \cite{Clip2protect_2023_CVPR} and conduct experiments on the CelebA-HQ dataset. 500 subjects are randomly selected, each with a pair of images. One image from each pair is assigned to the training set, while the other is designated for the test set. The selected images impersonate a synthesized target identity chosen from the “100k Faces Generated by AI” database\footnote{\url{https://github.com/ozgrozer/100k-faces}}. Quantitative and qualitative results are presented in Tab. \ref{tab_3} and Fig. \ref{fig_7}, respectively. 

\begin{table}[h]
\centering
\resizebox{0.8\columnwidth}{!}
{\begin{tabular}{c|cccc}
\hline
Method & IRSE50 & IR152 & Facenet & Mobileface  \\ \hline
CLIP2Protect  & 83.4  &  83.6  &  93.5  &  62.8 \\ \hline
\textbf{Ours} & 90.2  &  88.6  &  87.0  &  85.0 \\ \hline
\end{tabular}}
\caption{Protection success rate (PSR) under the black-box setting.}
\label{tab_3}
\end{table}

\section{Conclusion}
\label{sec:conclusion}
In this study, we explored the use of diffusion models for facial privacy protection, leveraging their powerful image-generation capabilities. We proposed a novel framework that adversarially modifies the latent diffusion model's latent code, rather than changing pixel space. We introduced learned unconditional embeddings as null-text guidance to weaken the diffusion model's purification effect and significantly enhanced privacy protection. Additionally, we incorporated self-attention guidance to preserve structural integrity between the original and generated images, ensuring high visual quality. Comprehensive quantitative and qualitative evaluations demonstrate the efficacy of our approach in protecting facial privacy compared to various FR models while maintaining superior visual quality.

\textbf{Acknowledgements.} This work was supported by the Research Council of Finland (former Academy of Finland) Academy Professor project EmotionAI (grants 336116, 345122, 359854), HPC project FaceCanvas (grant number 364905), the University of Oulu \& Research Council of Finland Profi 7 (grant 352788), EU HORIZON-MSCA-SE-2022 project ACMod (grant 101130271), Academy Research Fellow project (grant 355095), and Infotech Oulu. The authors also wish to acknowledge CSC – IT Center for Science, Finland, for computational resources.
{
    \small
    \bibliographystyle{ieeenat_fullname}
    \bibliography{main}
}

\clearpage
\setcounter{page}{1}
\maketitlesupplementary

In this supplementary, we first review diffusion models in Sec. \ref{sec:bkg} as they form the foundation of our proposed framework. Next, in Sec. \ref{sec:t_img} and Sec. \ref{sec:f_model}, we introduce the four target identities and face recognition (FR) models used in our experiments. Sec. \ref{sec:param_set} describes the weight factor of adversarial loss. Then, we further assess the effectiveness of our approach via some ablation studies in Sec. \ref{sec:ablation}. Finally, we present additional visualization results for a more comprehensive assessment in Sec. \ref{sec:vis}.

\section{Background:  Latent Diffusion Model}
\label{sec:bkg}
Diffusion models \cite{DDPM_2020_ANIPS, DDIM_2021_ICLR, Stable_diff_2022_CVPR} consist of two processes: (1) a T-step forward diffusion process that progressively corrupts the input image $x$ with Gaussian noise until it approaches a Gaussian distribution $x_T$ at step T;  (2) a reverse denoising process, which seeks to recover $x$ from $x_T$ by gradual reducing noise over T reverse steps. Unlike the denoising diffusion probabilistic model (DDPM) \cite{DDPM_2020_ANIPS}, the latent diffusion model (LDM) \cite{Stable_diff_2022_CVPR} operates in the latent rather than pixel space.  In LDM, an autoencoder first compresses the image into a lower-dimensional latent representation $z$. The diffusion process then applies noise and denoising within this latent space. Finally, the latent representation is decoded back to the original image space. The forward process in LDM is defined as:
\begin{equation}
    \label{eq8}
    q\left(z_{t} \mid z_{t-1}\right)=\mathcal{N}\left(z_{t} ; \sqrt{1-\beta_{t}} z_{t-1}, \beta_{t} \mathbf{I}\right)\;,
\end{equation}
where $\beta_{t} \in(0,1]$ are parameters control the noise level at each diffusion step $t$. An important property of the forward process is that $z_{t}$ can be directly sampled at any time $t$ given the original latent variable $z_{0}$ using:
\begin{equation}
    \label{eq9}
    q\left(z_{t} \mid z_{0}\right)=\mathcal{N}\left(z_{t} ; \sqrt{\bar{\alpha}_{t}} z_{0},\left(1-\bar{\alpha}_{t}\right) \mathbf{I}\right)\;,
\end{equation}
where $\bar{\alpha}_{t}=\prod_{s=1}^{t}\left(1-\beta_{s}\right)$. Given that the reverse process $q\left(z_{t-1} \mid z_{t}\right)$ is intractable due to its dependence on the unknown data distribution $q\left(z_{0}\right)$, it can be approximated using a parameterized Gaussian transition model conditioned on a context $\mathcal{C}$, which is formulated as follows:
\begin{multline}
    \label{eq10}
    p_{\theta}\left(z_{t-1} \mid z_{t}, \mathcal{C}\right)=
    \\\mathcal{N}\left(z_{t-1} ; \mu_{\theta}\left(z_{t}, t, \mathcal{C}\right), \Sigma_{\theta}\left(z_{t}, t, \mathcal{C}\right)\right)\;,
\end{multline}
where $\mu_{\theta}$ and $\Sigma_{\theta}$ are mean and covariance matrix. The mean $\mu_{\theta}$ can be expressed as:
\begin{equation}
    \label{eq11}
    \mu_{\theta}\left(z_{t}, t, \mathcal{C}\right)=\frac{1}{\sqrt{\alpha_{t}}}\left(z_{t}-\frac{\beta_{t}}{\sqrt{1-\bar{\alpha}_{t}}} \epsilon_{\theta}\left(z_{t}, t, \mathcal{C}\right)\right)\;.
\end{equation}
Here $\epsilon_{\theta}\left(z_{t}, t, \mathcal{C}\right)$ is the model's prediction of the noise added at time step $t$, given the conditioning information $\mathcal{C}$. After training the model $\epsilon_{\theta}\left(z_{t}, t, \mathcal{C}\right)$, the following sampling method can be employed:
\begin{equation}
    \label{eq12}
    z_{t-1}=\mu_{\theta}\left(z_{t}, t, \mathcal{C}\right)+\sigma_{t} z, \quad z \sim \mathcal{N}(0,1)\;.
\end{equation}
To accelerate image generation, Song \etal. \cite{DDIM_2021_ICLR} introduce the denoising diffusion implicit model (DDIM), which employs a non-Markovian reverse process, as shown below:
\begin{multline}
    \label{eq13}
    z_{t-1}=\sqrt{\bar{\alpha}_{t-1} / \bar{\alpha}_{t}} z_{t} + \\\sqrt{\bar{\alpha}_{t-1}}\left(\sqrt{1 / \bar{\alpha}_{t-1}-1}-\sqrt{1 / \bar{\alpha}_{t}-1}\right) \epsilon_{\theta}\left(z_{t}, t, \mathcal{C}\right)\;.
\end{multline}
Equation \ref{eq13} is derived from Equation \ref{eq12} by eliminating the stochastic noise component ($\sigma_{t}=0$), following the DDIM's principle, and substituting it with a deterministic process.

\section{Target Images}
\label{sec:t_img}
\begin{figure}[t]
    \centering
    \resizebox{\columnwidth}{!}
    {\begin{tabular}{cccc}
        \begin{subfigure}[t]{0.2\textwidth}
            \includegraphics[width=\linewidth]{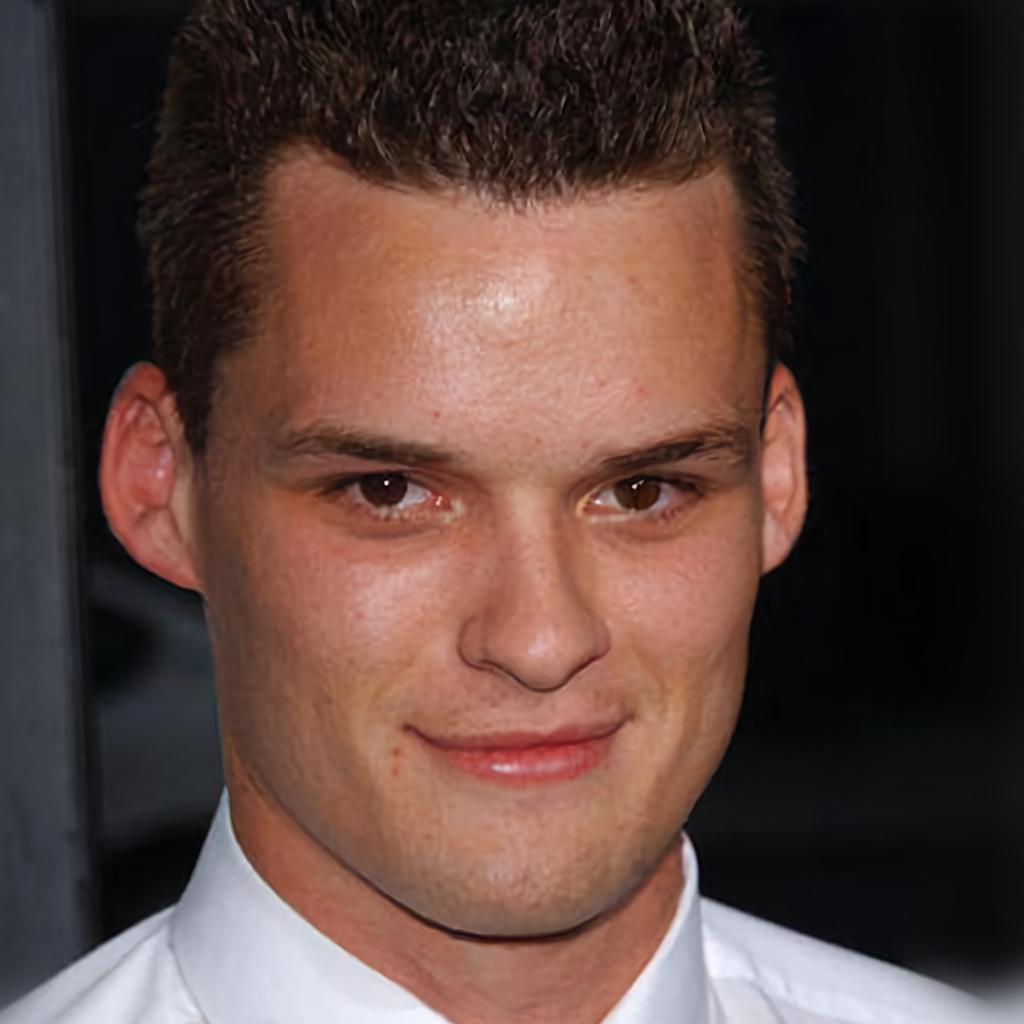}
            \centering
            \vspace{2pt}
            \includegraphics[width=\linewidth]{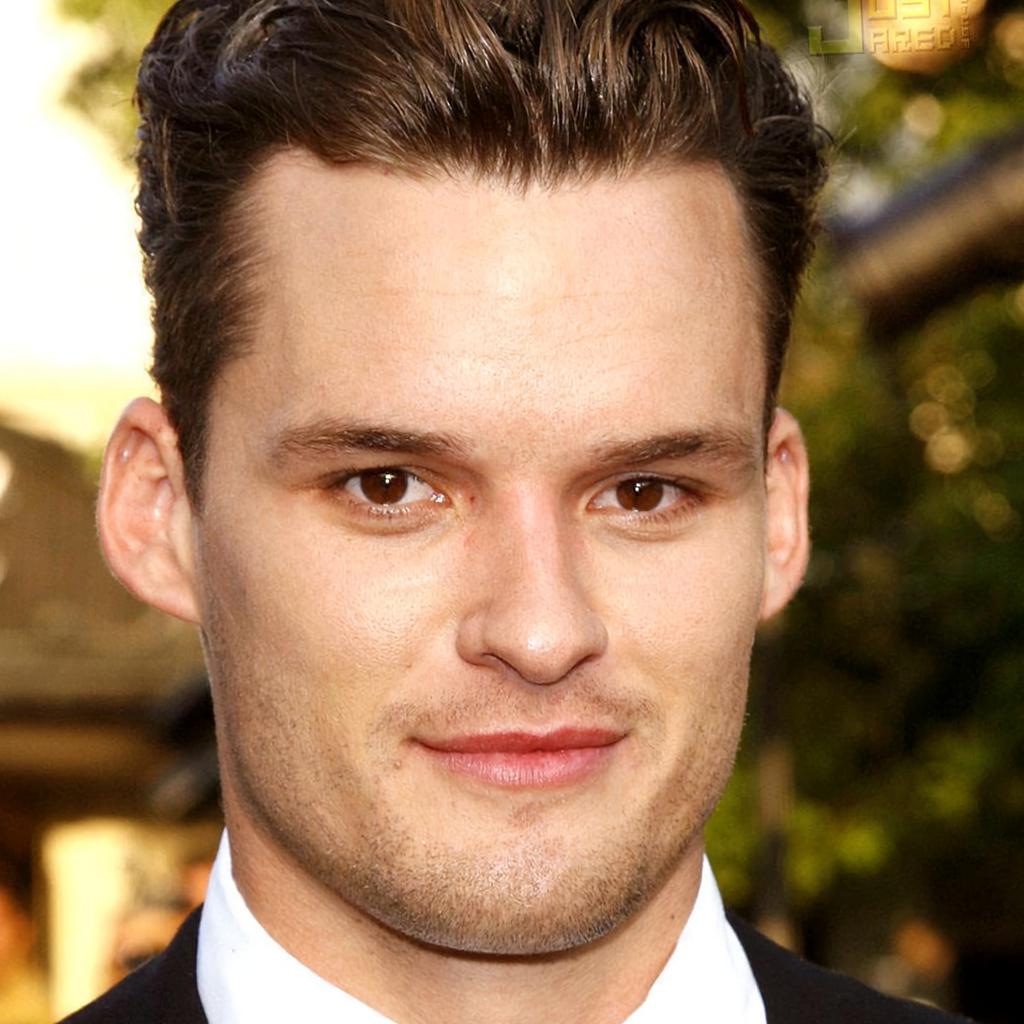}
            \vspace{2pt}
            \fontsize{12pt}{14pt}\selectfont{Target 1}
        \end{subfigure}
        \begin{subfigure}[t]{0.2\textwidth}
            \includegraphics[width=\linewidth]{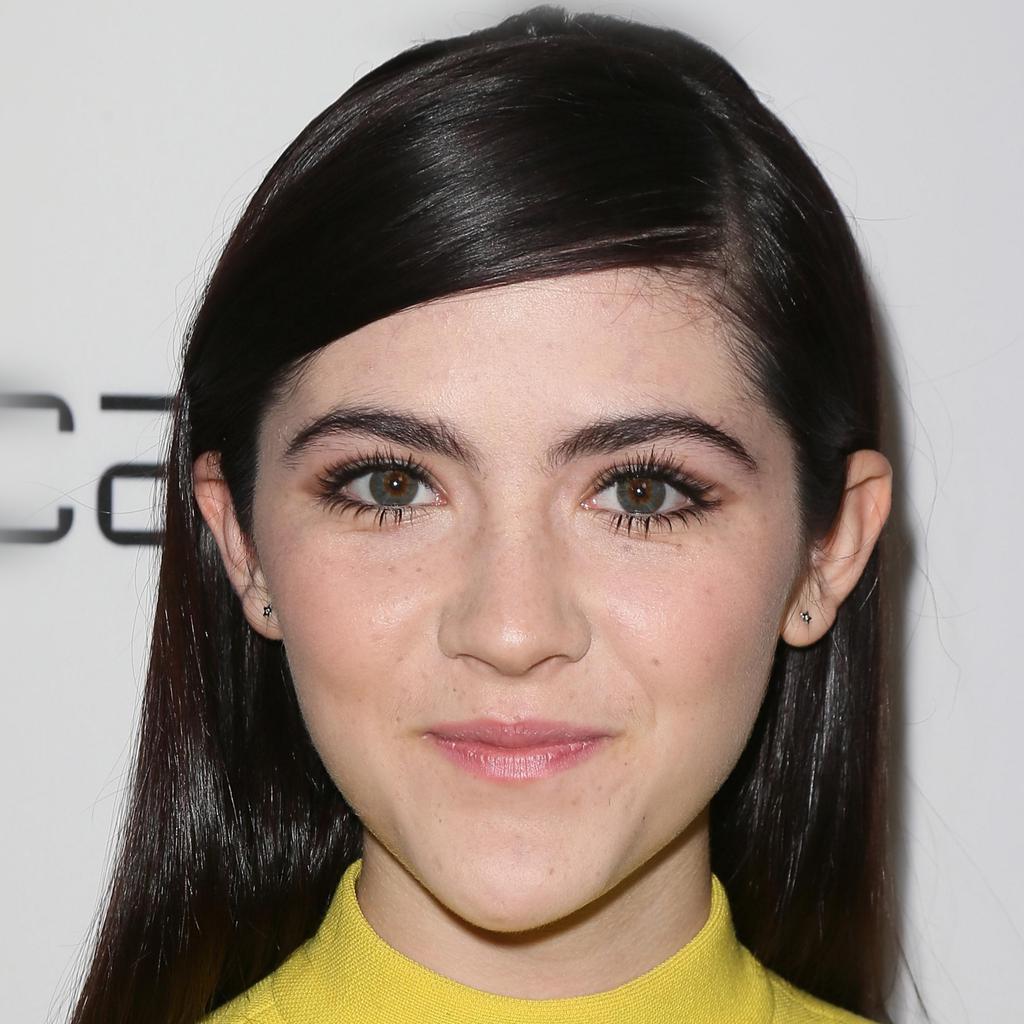}
            \centering
            \vspace{2pt}
            \includegraphics[width=\linewidth]{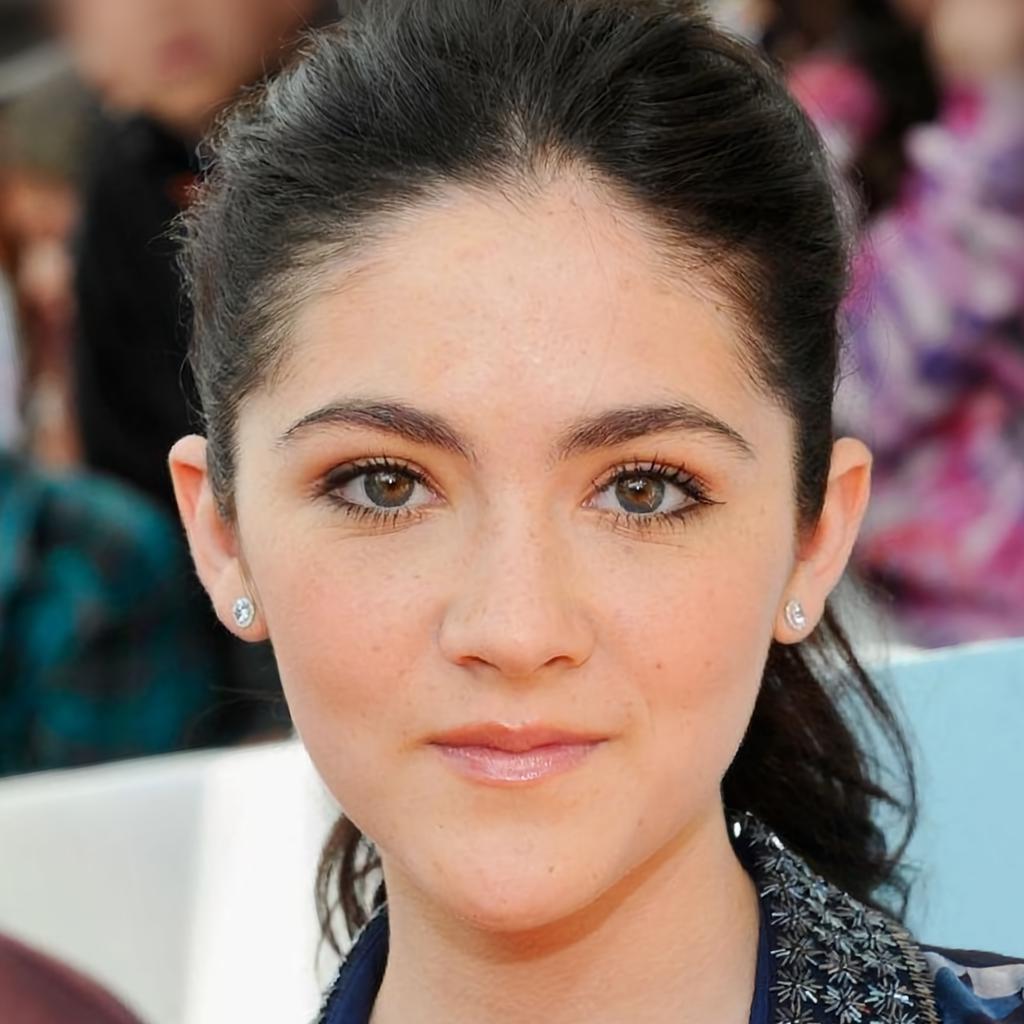}
            \vspace{2pt}
            \fontsize{12pt}{14pt}\selectfont{Target 2}
        \end{subfigure}
        \begin{subfigure}[t]{0.2\textwidth}
            \includegraphics[width=\linewidth]{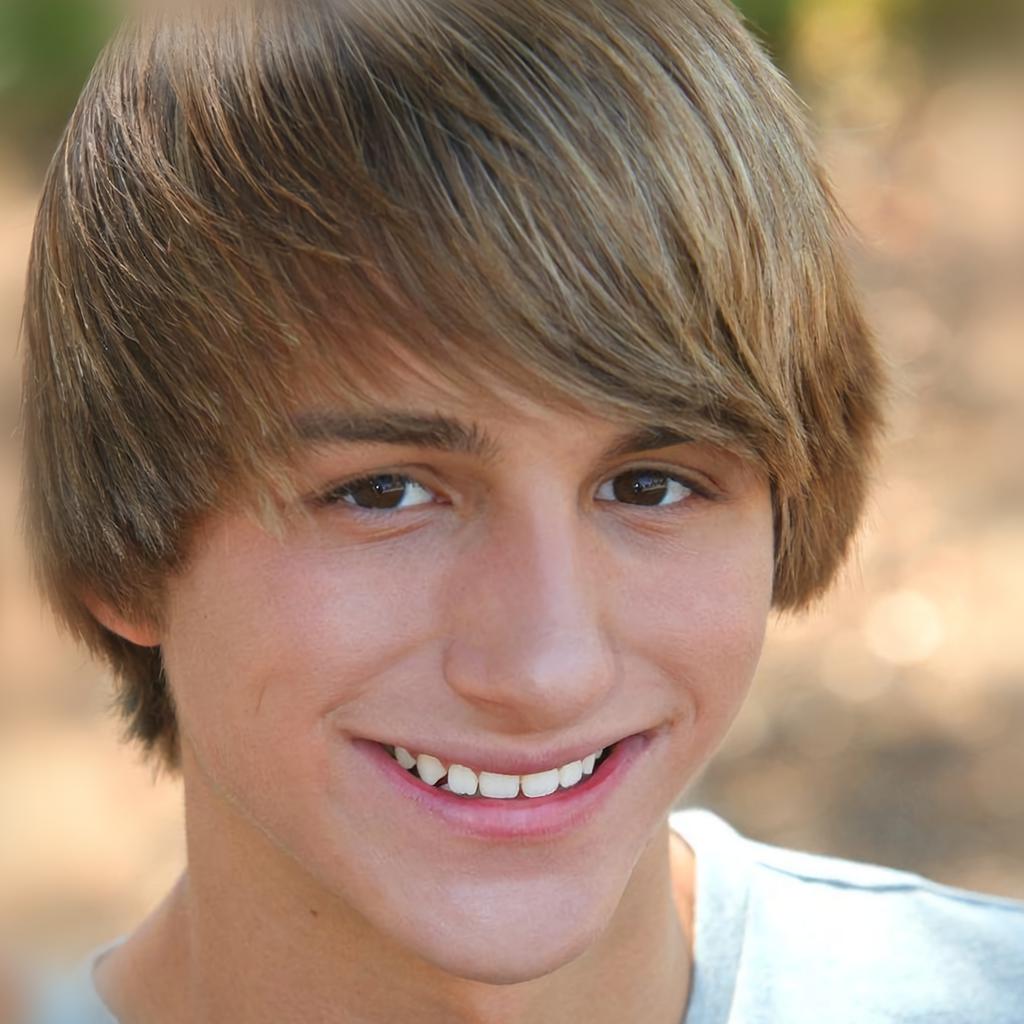}
            \centering
            \vspace{2pt}
            \includegraphics[width=\linewidth]{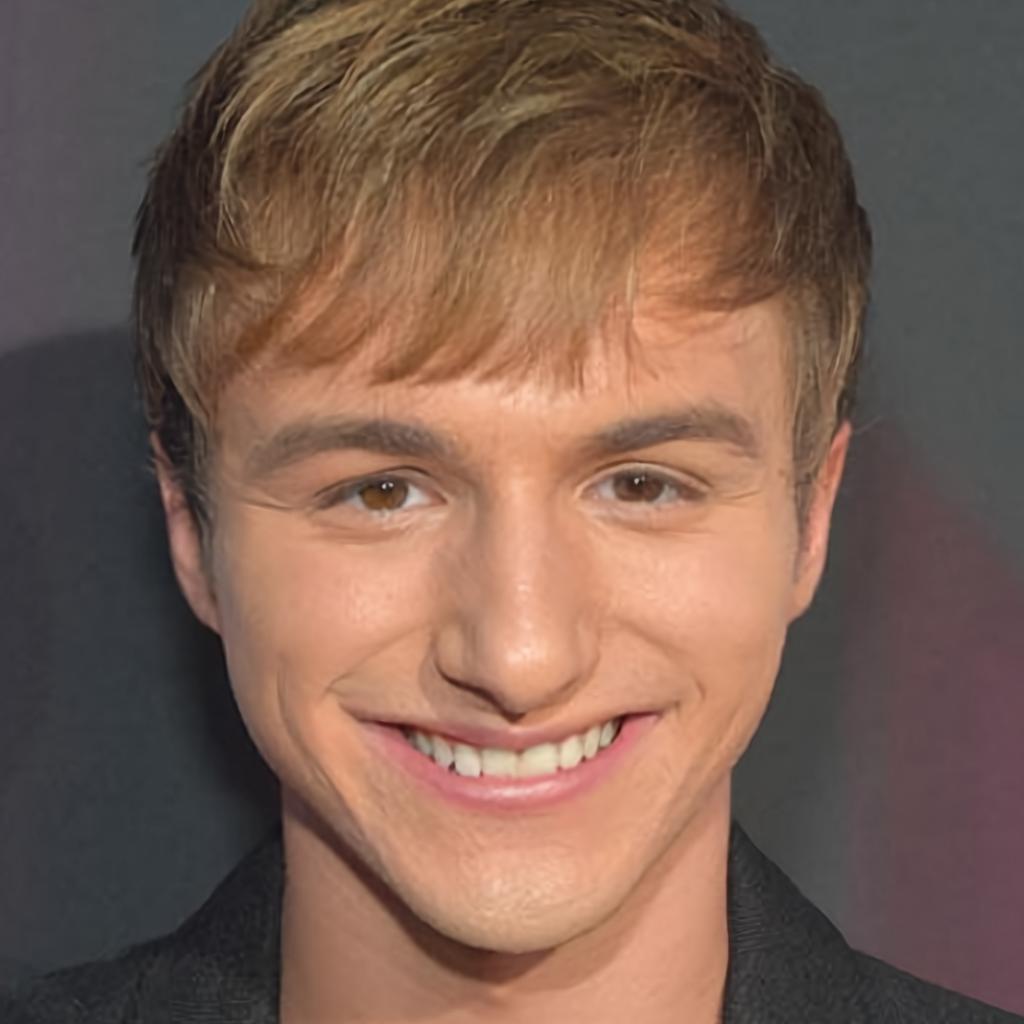}
            \vspace{2pt}
            \fontsize{12pt}{14pt}\selectfont{Target 3}
        \end{subfigure}
        \begin{subfigure}[t]{0.2\textwidth}
            \includegraphics[width=\linewidth]{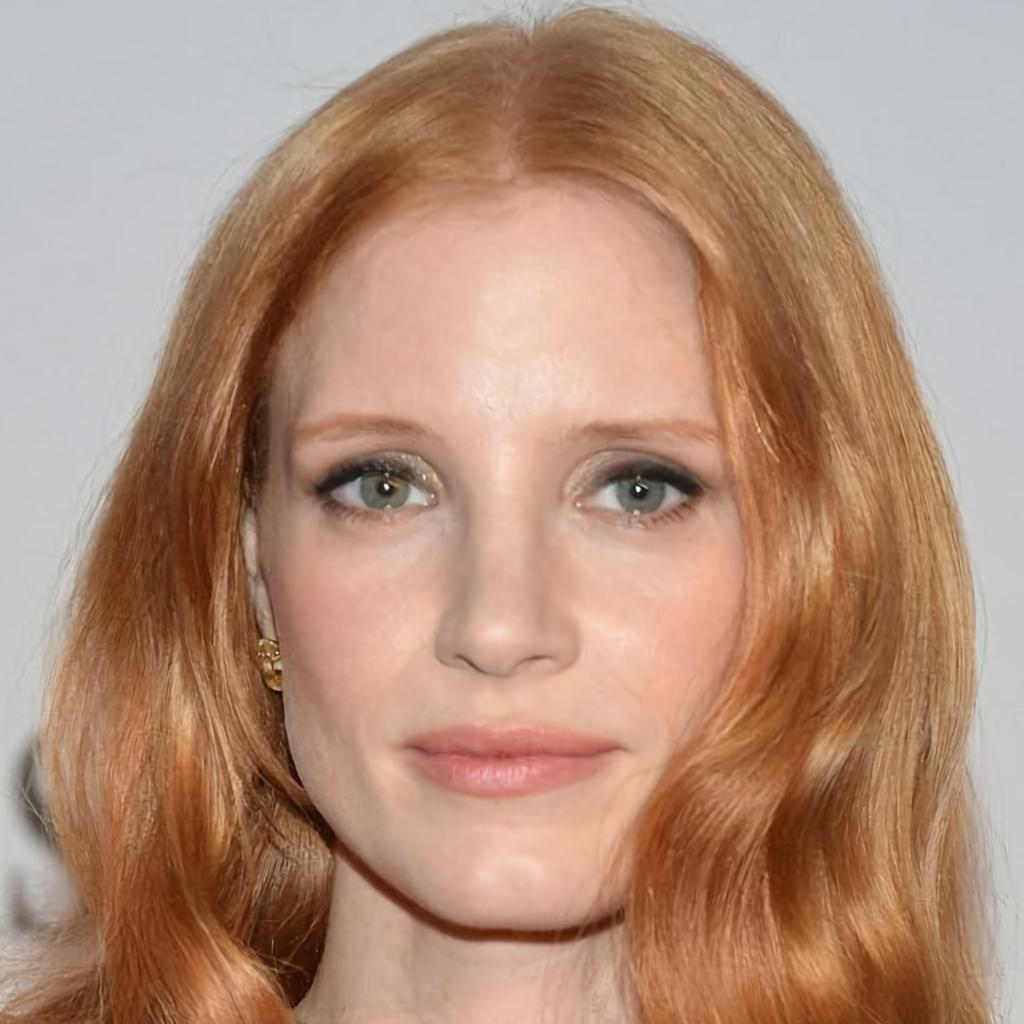}
            \centering
            \vspace{2pt}
            \includegraphics[width=\linewidth]{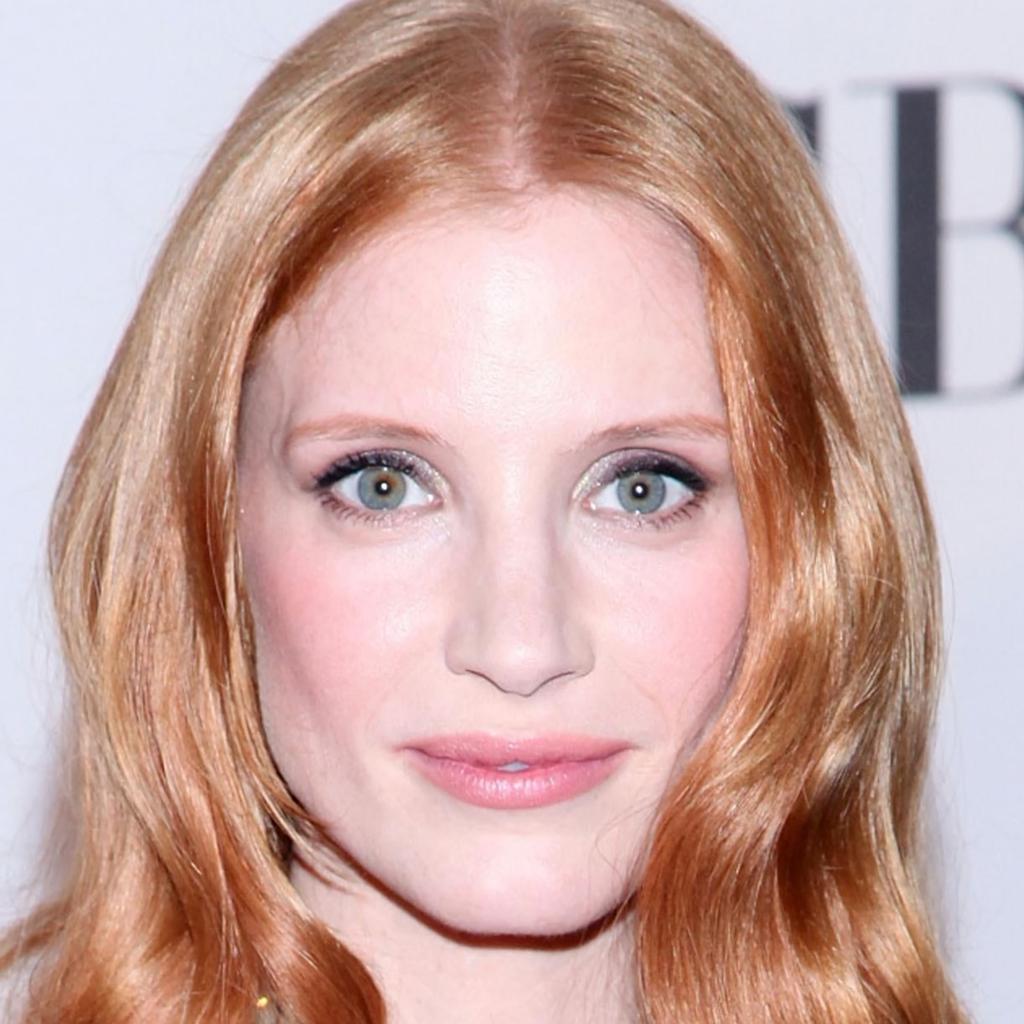}
            \vspace{2pt}
            \fontsize{12pt}{14pt}\selectfont{Target 4}
        \end{subfigure}
    \end{tabular}}
    \caption{Target identities used for impersonation. The first row contains images used for training, while the second includes images used for testing.}
    \label{fig_8}
\end{figure}

The proposed model is designed to generate protected face images that deceive malicious FR models into misidentifying those protected faces as a specified target identity. Fig. \ref{fig_8} presents the four target identities provided by \cite{AMT-GAN_2022_CVPR} mentioned in the Experiments Section (Sec. 4 in the main paper). To better mimic real-world scenarios, we ensure that the target images used during training differ from those used during testing.

\section{Face Recognition Models}
\label{sec:f_model}
For fair comparisons, we adopt publicly available pre-trained FR models following \cite{AMT-GAN_2022_CVPR}. Three of these models are based on ArcFace \cite{Mobileface_2019_CVPR}, the state-of-the-art FR algorithm, which processes facial images at a resolution of $112\times 112$ and encodes them into 512-dimensional feature vectors. These models differ in their neural architectures and training datasets: IR152 \cite{IR152_2016_CVPR} employs ResNet-152, IRSE50 \cite{IRSE50_2018_CVPR} uses ResNet-50, and MobileFace \cite{Mobilefacenets_2018} is built on MobileFaceNet. Facenet \cite{Facenet_2015_CVPR}, on the other hand, leverages InceptionResnet \cite{inception_2017_AAAI} and follows the original training protocols outlined in its paper, using an input resolution of $160\times 160$. To assess the models' effectiveness, we report their FR accuracy on the CelebA-HQ dataset: IR152: 90.70\%, IRSE50: 90.80\%, MobileFace: 83.00\%, and Facenet: 91.20\%.

\section{Parameter Settings}
\label{sec:param_set}
\begin{figure}[!t]
    \centering
    \resizebox{\columnwidth}{!}
    {\begin{tikzpicture}
        \begin{axis}[
        axis y line*=left,
        ymin=88, ymax=92,
        xlabel=$\lambda_{adv}$,
        ylabel=PSR,
        xtick={0.001, 0.003, 0.006, 0.01},
        xticklabels={0.001, 0.003, 0.006, 0.01},
        ytick={88, 89, 90, 91, 92},
        grid=major,
        width=10cm,
        height=6cm,
        ytick style={blue},
        yticklabel style=blue,
        mark size=2pt,
        scaled x ticks=false,
        ]
        \addplot+[blue, mark=*, solid, mark options={solid}, line width=1pt] coordinates {
            (0.001, 89.06)
            (0.003, 91.57)
            (0.006, 91.67)
            (0.010, 91.70)
        };
        \label{plot1}
        \end{axis}
    
        \begin{axis}[
            axis y line*=right,
            axis x line=none,
            ymin=11, ymax=15,
            ylabel=FID,
            ytick={11, 12, 13, 14, 15},
            width=10cm,
            height=6cm,
            ytick style={red},
            yticklabel style=red,
            mark size=2pt,
            legend pos=south east,
            legend style={nodes={scale=0.75, transform shape}}
        ]
        \addlegendimage{refstyle=plot1} \addlegendentry{PSR}
        \addlegendentry{FID}
        
        \addplot+[red, mark=*, solid, mark options={solid}, line width=1pt] coordinates {
            (0.001, 11.33)
            (0.003, 12.72)
            (0.006, 14.0)
            (0.010, 14.6)
        };
        \addlegendentry{FID}
        \end{axis}
    \end{tikzpicture}}
    \caption{Quantitative study on the parameter settings of the weight factor of adversarial loss.}
    \label{fig_9}
\end{figure}
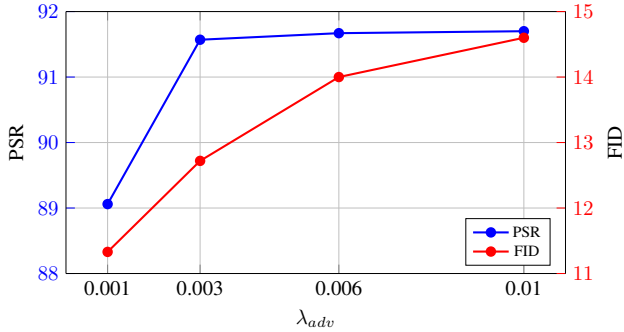
We evaluated the impact of varying the adversarial loss weight, $\lambda_{adv}$, on both privacy protection and image quality in Fig. \ref{fig_9}. The results indicate that increasing $\lambda_{adv}$ slightly improves privacy protection performance, as reflected by higher protection success rate (PSR). However, this comes at the expense of image quality, as evidenced by deteriorating Fréchet inception distance (FID). Conversely, lower $\lambda_{adv}$ yields better image quality but significantly weaker privacy protection.

\section{Ablation Studies}
\label{sec:ablation}
\subsection{Optimizing Latent Codes across Timesteps}
As mentioned in Sec. 3.4 in the main paper, the denoising model projects the perturbed noise back toward the natural data manifold during the reverse diffusion process. One potential solution to prevent the purification effect could be considering $z_i$ from all timesteps as the latent code, optimizing it throughout the adversarial latent code learning process. Instead of utilizing the learned unconditional embedding proposed in our approach, we conducted an additional experiment by optimizing $z_i$ across multiple timesteps from $t$ to 0. The results show that the PSR improves from 91.57 to 93.37 when optimizing from $z_3$ to $z_1$, comparable to and slightly better than our method. However, this improvement comes at the cost of image quality, as indicated by the increase in FID from 12.72 to 15.71, suggesting that the generated images exhibit more structural changes. Additionally, the computational complexity increases significantly, with generation time rising from 15 seconds when optimizing only $z_3$, to 23 seconds when optimizing from $z_3$ to $z_1$, and up to 40 seconds when optimizing from $z_5$ to $z_1$ (Experiments were conducted using MobileFace as the target model).\\
The comparable performance of null-text guidance indicates that it implicitly approximates the impact of optimizing the latent codes at different timesteps while offering substantial benefits in preserving image quality and computational efficiency.

\subsection{Effectiveness Against Adaptive Adversaries}
\begin{table}[t!]
\centering
\resizebox{0.9\columnwidth}{!}
{\begin{tabular}{c|cccc}
\hline
 & IRSE50 & IR152 & Facenet & Mobileface  \\ \hline
Ours w/o smoothing & 88.87  &  67.25  &  59.53  &  91.57 \\ \hline
Gauss$_{3\times 3}$ & 88.47 &  67.20  &  59.23  &  91.47 \\ \hline
Gauss$_{5\times 5}$ & 87.61  &  66.73  &  58.73  &  90.26 \\ \hline
Gauss$_{7\times 7}$ & 87.06  &  66.35  &  57.93  &  88.56 \\ \hline
Mean$_{5\times 5}$ & 86.66  &  65.75  &  57.33  & 87.86 \\ \hline
\end{tabular}}
\vspace{-0.2cm}
\caption{Protection success rate (PSR) of our method against adaptive adversaries.}
\label{tab_4}
\end{table}
An adaptive privacy adversary with advanced knowledge may deploy additional mechanisms to bypass the protection method. To evaluate the resilience of our approach under such adaptive scenarios, we assess its effectiveness against common image-smoothing techniques. Table \ref{tab_4} presents the results of applying Gaussian filters with kernel sizes of $3\times 3$, $5\times 5$, and $7\times 7$, as well as a mean filter with a $5\times 5$ kernel—widely used methods in the adversarial robustness domain. Despite slight degradation, PSR remains relatively high after smoothing, indicating that our approach maintains robust protection against these countermeasures.

\subsection{Protection Performances on Commercial APIs}
\label{sec:commercial}
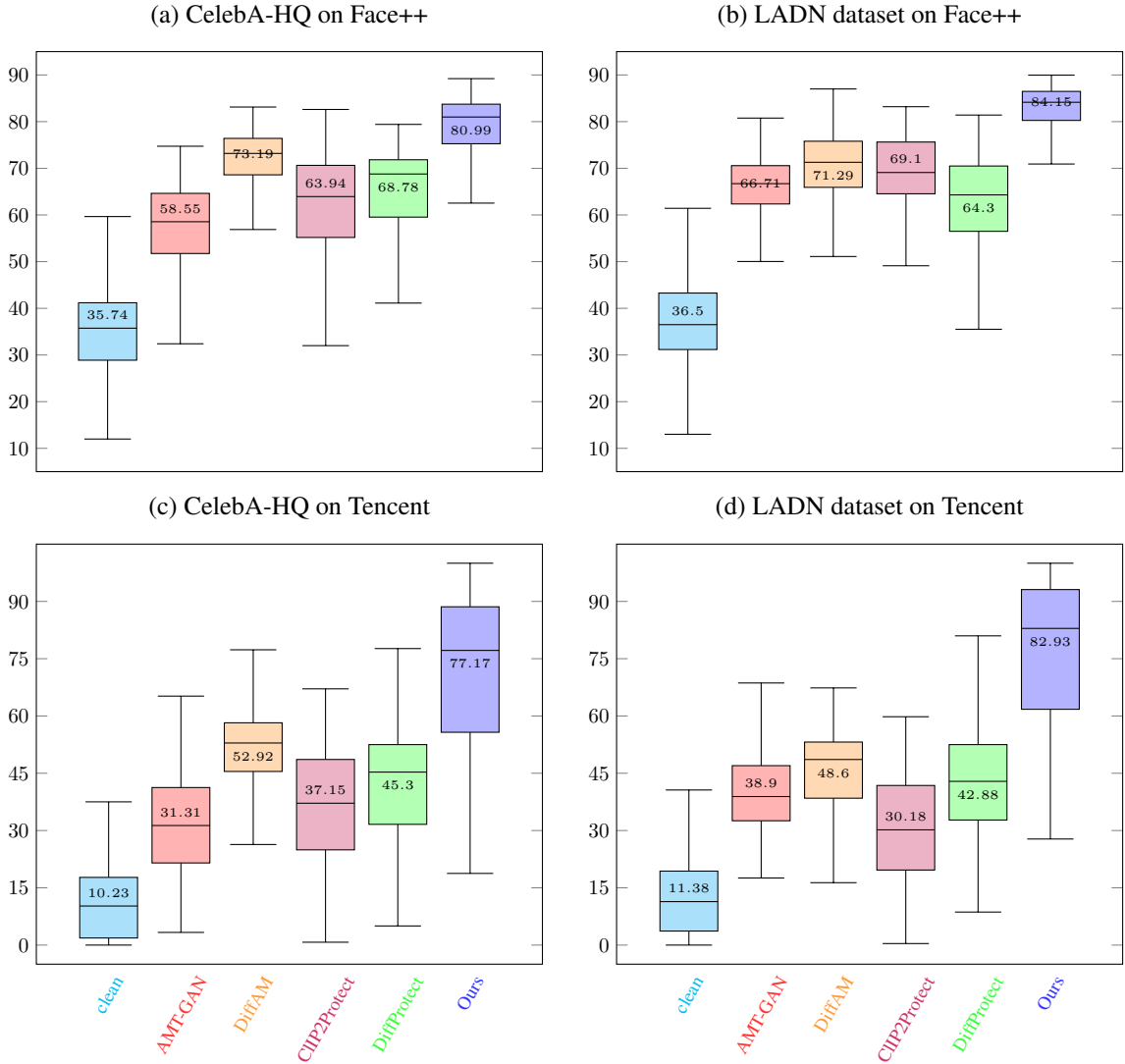
\begin{figure*}[t]
    \centering
    \pgfplotsset{compat=1.18}
    \usepgfplotslibrary{statistics}
    \begin{minipage}{0.45\textwidth}
        \centering
        \begin{tikzpicture}
            \begin{axis}[
                boxplot/draw direction = y,
    		xticklabels = {draw=none},
    		xtick style = {draw=none}, 
    		title={(a) CelebA-HQ on Face++},
                ymin=5, ymax=95,
                ytick = {10, 20, 30, 40, 50, 60, 70, 80, 90},
                ticklabel style={scale=0.8},
                cycle list={{black}},
            ]
            \addplot + [boxplot prepared={
                lower whisker=11.99, lower quartile=28.88,
                median=35.74, upper quartile=41.19,
                upper whisker=59.66},
                draw=black,
                fill=cyan!30,
            ] coordinates {}
            [above, font=\tiny, color=black]
            node at (boxplot box cs: \boxplotvalue{median}, 0.5) {\pgfmathprintnumber{\boxplotvalue{median}}};
            \addplot + [boxplot prepared={
                lower whisker=32.40, lower quartile=51.74,
                median=58.55, upper quartile=64.63,
                upper whisker=74.73},
                draw=black,
                fill=red!30,
            ] coordinates {}
            [above, font=\tiny, color=black]
            node at (boxplot box cs: \boxplotvalue{median}, 0.5) {\pgfmathprintnumber{\boxplotvalue{median}}};
            \addplot + [boxplot prepared={
                lower whisker=56.88, lower quartile=68.60,
                median=73.19, upper quartile=76.41,
                upper whisker=83.13},
                draw=black,
                fill=orange!30,
            ] coordinates {}
            [font=\tiny, color=black]
            node at (boxplot box cs: \boxplotvalue{median}, 0.5) {\pgfmathprintnumber{\boxplotvalue{median}}};
            \addplot + [boxplot prepared={
                lower whisker=32.00, lower quartile=55.18,
                median=63.94, upper quartile=70.63,
                upper whisker=82.61},
                draw=black,
                fill=purple!30,
            ] coordinates {}
            [above, font=\tiny, color=black]
            node at (boxplot box cs: \boxplotvalue{median}, 0.5) {\pgfmathprintnumber{\boxplotvalue{median}}};
            \addplot + [boxplot prepared={
                lower whisker=41.12, lower quartile=59.54,
                median=68.78, upper quartile=71.83,
                upper whisker=79.41},
                draw=black,
                fill=green!30,
            ] coordinates {}
            [below, font=\tiny, color=black]
            node at (boxplot box cs: \boxplotvalue{median}, 0.5) {\pgfmathprintnumber{\boxplotvalue{median}}};
            \addplot + [boxplot prepared={
                lower whisker=62.54, lower quartile=75.26,
                median=80.99, upper quartile=83.74,
                upper whisker=89.21},
                draw=black,
                fill=blue!30,
            ] coordinates {}
            [below, font=\tiny, color=black]
            node at (boxplot box cs: \boxplotvalue{median}, 0.5) {\pgfmathprintnumber{\boxplotvalue{median}}};
            \end{axis}
        \end{tikzpicture}
    \end{minipage}%
    \begin{minipage}{0.45\textwidth}
        \centering
        \begin{tikzpicture}
            \begin{axis}[
                boxplot/draw direction = y,
                xticklabels = {draw=none},
    		xtick style = {draw=none}, 
    		title={(b) LADN dataset on Face++},
                ymin=5, ymax=95,
                ytick = {10, 20, 30, 40, 50, 60, 70, 80, 90},
                ticklabel style={scale=0.8},
                cycle list={{black}},
            ]
            \addplot + [boxplot prepared={
                lower whisker=13.01, lower quartile=31.17,
                median=36.50, upper quartile=43.27,
                upper whisker=61.43},
                draw=black,
                fill=cyan!30,
            ] coordinates {}
            [above, font=\tiny, color=black]
            node at (boxplot box cs: \boxplotvalue{median}, 0.5) {\pgfmathprintnumber{\boxplotvalue{median}}};
            \addplot + [boxplot prepared={
                lower whisker=50.03, lower quartile=62.36,
                median=66.71, upper quartile=70.57,
                upper whisker=80.76},
                draw=black,
                fill=red!30,
            ] coordinates {}
            [font=\tiny, color=black]
            node at (boxplot box cs: \boxplotvalue{median}, 0.5) {\pgfmathprintnumber{\boxplotvalue{median}}};
            \addplot + [boxplot prepared={
                lower whisker=51.10, lower quartile=65.92,
                median=71.29, upper quartile=75.80,
                upper whisker=87.00},
                draw=black,
                fill=orange!30,
            ] coordinates {}
            [below, font=\tiny, color=black]
            node at (boxplot box cs: \boxplotvalue{median}, 0.5) {\pgfmathprintnumber{\boxplotvalue{median}}};
            \addplot + [boxplot prepared={
                lower whisker=49.12, lower quartile=64.54,
                median=69.10, upper quartile=75.63,
                upper whisker=83.20},
                draw=black,
                fill=purple!30,
            ] coordinates {}
            [above, font=\tiny, color=black]
            node at (boxplot box cs: \boxplotvalue{median}, 0.5) {\pgfmathprintnumber{\boxplotvalue{median}}};
            \addplot + [boxplot prepared={
                lower whisker=35.51, lower quartile=56.50,
                median=64.30, upper quartile=70.49,
                upper whisker=81.41},
                draw=black,
                fill=green!30,
            ] coordinates {}
            [below, font=\tiny, color=black]
            node at (boxplot box cs: \boxplotvalue{median}, 0.5) {\pgfmathprintnumber{\boxplotvalue{median}}};
            \addplot + [boxplot prepared={
                lower whisker=70.92, lower quartile=80.26,
                median=84.15, upper quartile=86.48,
                upper whisker=89.96},
                draw=black,
                fill=blue!30,
            ] coordinates {}
            [font=\tiny, color=black]
            node at (boxplot box cs: \boxplotvalue{median}, 0.5) {\pgfmathprintnumber{\boxplotvalue{median}}};
            \end{axis}
        \end{tikzpicture}
    \end{minipage}%
    \\
    \begin{minipage}{0.45\textwidth}
        \centering
        \begin{tikzpicture}
            \begin{axis}[
                boxplot/draw direction = y,
                xmin=0, xmax=7,
                xtick = {1, 2, 3, 4, 5, 6},
    		xticklabels = {\color{cyan!80} clean, \color{red!80} AMT-GAN, \color{orange!80} DiffAM, \color{purple!80} ClIP2Protect, \color{green!80} DiffProtect, \color{blue!80} Ours},
                xticklabel style = {align=center, font=\small, rotate=60},
    		xtick style = {draw=none}, 
    		title={(c) CelebA-HQ on Tencent},
                ymin=-5, ymax=105,
                ytick = {0, 15, 30, 45, 60, 75, 90},
                ticklabel style={scale=0.8},
                cycle list={{black}},
            ]
            \addplot + [boxplot prepared={
                lower whisker=0, lower quartile=1.86,
                median=10.23, upper quartile=17.73,
                upper whisker=37.51},
                draw=black,
                fill=cyan!30,
            ] coordinates {}
            [above, font=\tiny, color=black]
            node at (boxplot box cs: \boxplotvalue{median}, 0.5) {\pgfmathprintnumber{\boxplotvalue{median}}};
            \addplot + [boxplot prepared={
                lower whisker=3.32, lower quartile=21.51,
                median=31.31, upper quartile=41.26,
                upper whisker=65.20},
                draw=black,
                fill=red!30,
            ] coordinates {}
            [above, font=\tiny, color=black]
            node at (boxplot box cs: \boxplotvalue{median}, 0.5) {\pgfmathprintnumber{\boxplotvalue{median}}};
            \addplot + [boxplot prepared={
                lower whisker=26.34, lower quartile=45.47,
                median=52.92, upper quartile=58.21,
                upper whisker=77.33},
                draw=black,
                fill=orange!30,
            ] coordinates {}
            [below, font=\tiny, color=black]
            node at (boxplot box cs: \boxplotvalue{median}, 0.5) {\pgfmathprintnumber{\boxplotvalue{median}}};
            \addplot + [boxplot prepared={
                lower whisker=0.75, lower quartile=24.94,
                median=37.15, upper quartile=48.61,
                upper whisker=67.10},
                draw=black,
                fill=purple!30,
            ] coordinates {}
            [above, font=\tiny, color=black]
            node at (boxplot box cs: \boxplotvalue{median}, 0.5) {\pgfmathprintnumber{\boxplotvalue{median}}};
            \addplot + [boxplot prepared={
                lower whisker=5.01, lower quartile=31.62,
                median=45.30, upper quartile=52.50,
                upper whisker=77.65},
                draw=black,
                fill=green!30,
            ] coordinates {}
            [below, font=\tiny, color=black]
            node at (boxplot box cs: \boxplotvalue{median}, 0.5) {\pgfmathprintnumber{\boxplotvalue{median}}};
            \addplot + [boxplot prepared={
                lower whisker=18.77, lower quartile=55.73,
                median=77.17, upper quartile=88.59,
                upper whisker=100},
                draw=black,
                fill=blue!30,
            ] coordinates {}
            [below, font=\tiny, color=black]
            node at (boxplot box cs: \boxplotvalue{median}, 0.5) {\pgfmathprintnumber{\boxplotvalue{median}}};
            \end{axis}
        \end{tikzpicture}
    \end{minipage}%
    \begin{minipage}{0.45\textwidth}
        \centering
        \begin{tikzpicture}
            \begin{axis}[
                boxplot/draw direction = y,
                xmin=0, xmax=7,
                xtick = {1, 2, 3, 4, 5, 6},
    		xticklabels = {\color{cyan!80} clean, \color{red!80} AMT-GAN, \color{orange!80} DiffAM, \color{purple!80} ClIP2Protect, \color{green!80} DiffProtect, \color{blue!80} Ours},
                xticklabel style = {align=center, font=\small, rotate=60},
    		xtick style = {draw=none}, 
    		title={(d) LADN dataset on Tencent},
                ymin=-5, ymax=105,
                ytick = {0, 15, 30, 45, 60, 75, 90},
                ticklabel style={scale=0.8},
                cycle list={{black}},
            ]
            \addplot + [boxplot prepared={
                lower whisker=0, lower quartile=3.67,
                median=11.38, upper quartile=19.37,
                upper whisker=40.62},
                draw=black,
                fill=cyan!30,
            ] coordinates {}
            [above, font=\tiny, color=black]
            node at (boxplot box cs: \boxplotvalue{median}, 0.5) {\pgfmathprintnumber{\boxplotvalue{median}}};
            \addplot + [boxplot prepared={
                lower whisker=17.57, lower quartile=32.56,
                median=38.90, upper quartile=46.99,
                upper whisker=68.64},
                draw=black,
                fill=red!30,
            ] coordinates {}
            [above, font=\tiny, color=black]
            node at (boxplot box cs: \boxplotvalue{median}, 0.5) {\pgfmathprintnumber{\boxplotvalue{median}}};
            \addplot + [boxplot prepared={
                lower whisker=16.34, lower quartile=38.47,
                median=48.60, upper quartile=53.17,
                upper whisker=67.35},
                draw=black,
                fill=orange!30,
            ] coordinates {}
            [below, font=\tiny, color=black]
            node at (boxplot box cs: \boxplotvalue{median}, 0.5) {\pgfmathprintnumber{\boxplotvalue{median}}};
            \addplot + [boxplot prepared={
                lower whisker=0.4, lower quartile=19.65,
                median=30.18, upper quartile=41.80,
                upper whisker=59.80},
                draw=black,
                fill=purple!30,
            ] coordinates {}
            [above, font=\tiny, color=black]
            node at (boxplot box cs: \boxplotvalue{median}, 0.5) {\pgfmathprintnumber{\boxplotvalue{median}}};
            \addplot + [boxplot prepared={
                lower whisker=8.65, lower quartile=32.75,
                median=42.88, upper quartile=52.50,
                upper whisker=80.98},
                draw=black,
                fill=green!30,
            ] coordinates {}
            [below, font=\tiny, color=black]
            node at (boxplot box cs: \boxplotvalue{median}, 0.5) {\pgfmathprintnumber{\boxplotvalue{median}}};
            \addplot + [boxplot prepared={
                lower whisker=27.78, lower quartile=61.75,
                median=82.93, upper quartile=93.10,
                upper whisker=100},
                draw=black,
                fill=blue!30,
            ] coordinates {}
            [below, font=\tiny, color=black]
            node at (boxplot box cs: \boxplotvalue{median}, 0.5) {\pgfmathprintnumber{\boxplotvalue{median}}};
            \end{axis}
        \end{tikzpicture}
    \end{minipage}%
    \caption{The confidence scores returned from Face++ and Tencent APIs. The higher confidence score indicates better protection performance. Our approach has a higher confidence score compared to four state-of-the-art methods, i.e., AMT-GAN \cite{AMT-GAN_2022_CVPR}, DiffAM\cite{DiffAM_2024_CVPR}, ClIP2Protect\cite{Clip2protect_2023_CVPR}, and DiffProtect\cite{Diffprotect_2023_arXiv}.}
    \label{fig_10}
\end{figure*}

In Fig. \ref{fig_10}, we further evaluate the protection performance of our proposed approach alongside other benchmarks using two commercial FR APIs, i.e., Face++\footnote{\url{https://www.faceplusplus.com/face-comparing/}} and Tencent\footnote{\url{https://cloud.tencent.com/product/facerecognition}}, to simulate real-world conditions. We randomly select 100 images from the CelebA-HQ \cite{CelebA-HQ_2017_arXiv} and 100 images from LADN \cite{LADN_2019_ICCV} datasets for protection, recording the confidence scores returned by each API. These scores range from 0 to 100, with higher values indicating greater similarity between the protected image and the target identity. The results show that our method achieves the highest confidence score compared to other approaches.

\section{More Visualization Results}
\label{sec:vis}
\begin{figure*}[h]
    \centering
    \resizebox{\textwidth}{!}
    {\begin{tabular}{ccccccc}
        \begin{subfigure}[t]{0.2\textwidth}
            \includegraphics[width=\linewidth]{images/fig8/005869.jpg}
            \centering
            \vspace{2pt}
            \fontsize{12pt}{14pt}\selectfont{Target 1}
        \end{subfigure}
        \begin{subfigure}[t]{0.2\textwidth}
            \includegraphics[width=\linewidth]{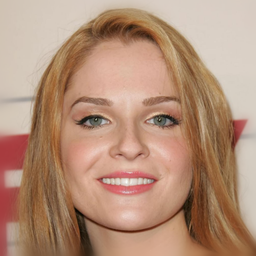}
            \centering
            \vspace{2pt}
            \includegraphics[width=\linewidth]{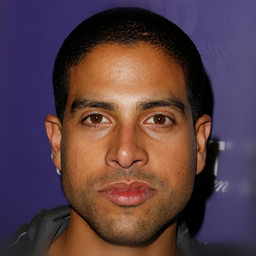}
            \centering
        \end{subfigure}
        \begin{subfigure}[t]{0.2\textwidth}
            \includegraphics[width=\linewidth]{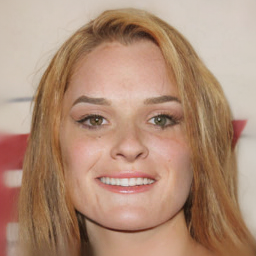}
            \centering
            \vspace{2pt}
            \includegraphics[width=\linewidth]{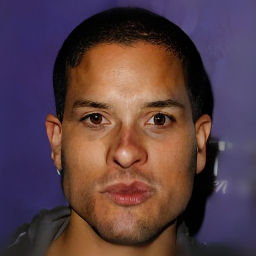}
            \centering
        \end{subfigure}
        \begin{subfigure}[t]{0.2\textwidth}
            \includegraphics[width=\linewidth]{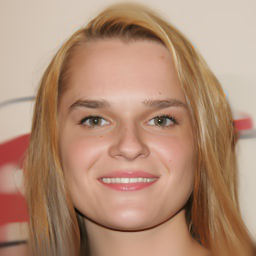}
            \centering
            \vspace{2pt}
            \includegraphics[width=\linewidth]{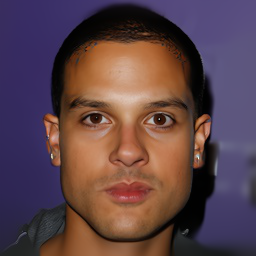}
            \centering
        \end{subfigure}
        \begin{subfigure}[t]{0.2\textwidth}
            \includegraphics[width=\linewidth]{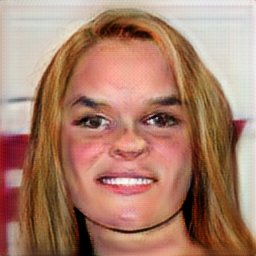}
            \centering
            \vspace{2pt}
            \includegraphics[width=\linewidth]{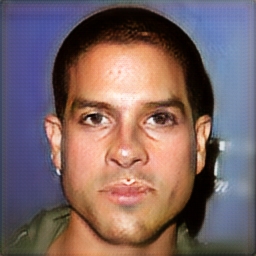}
            \centering
        \end{subfigure}
        \begin{subfigure}[t]{0.2\textwidth}
            \includegraphics[width=\linewidth]{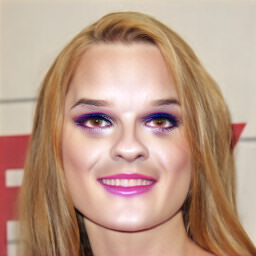}
            \centering
            \vspace{2pt}
            \includegraphics[width=\linewidth]{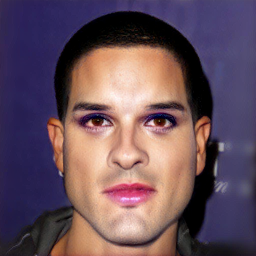}
            \centering
        \end{subfigure}
        \begin{subfigure}[t]{0.2\textwidth}
            \includegraphics[width=\linewidth]{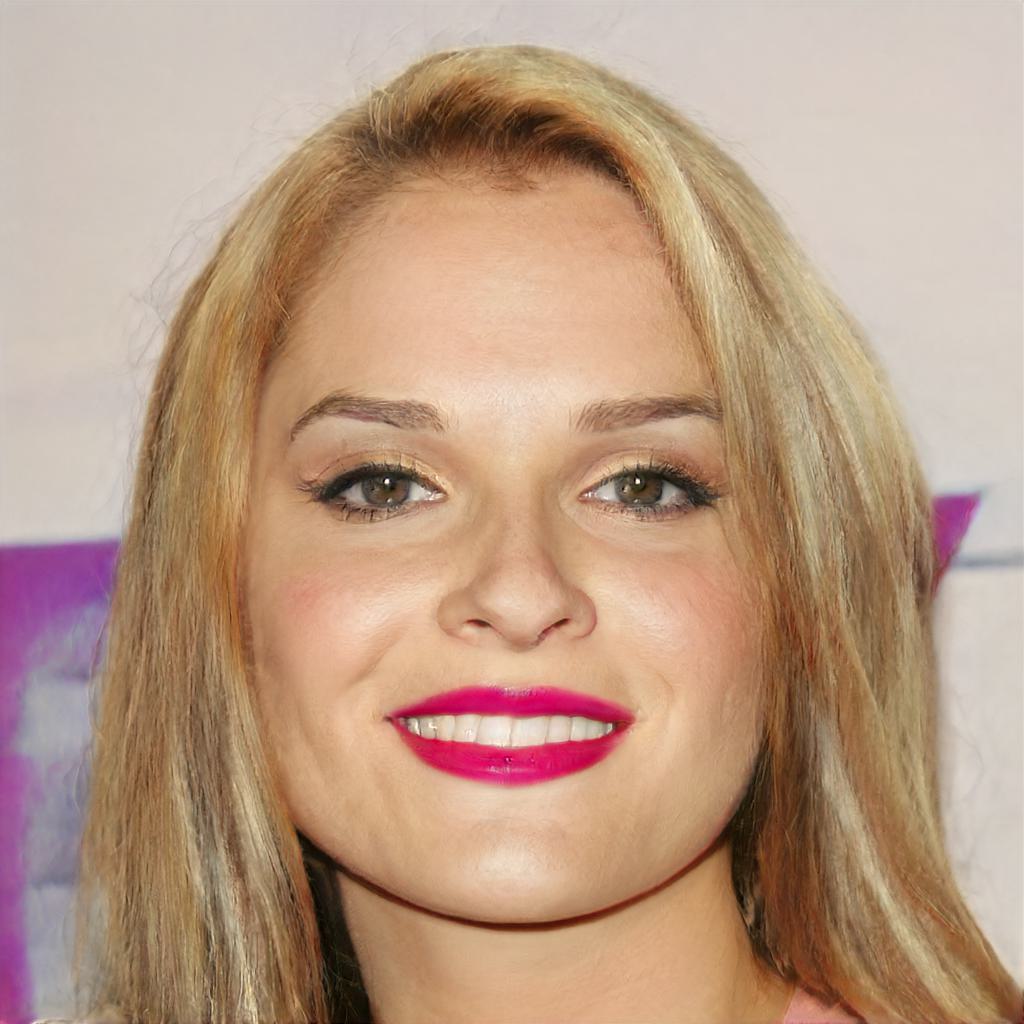}
            \centering
            \vspace{2pt}
            \includegraphics[width=\linewidth]{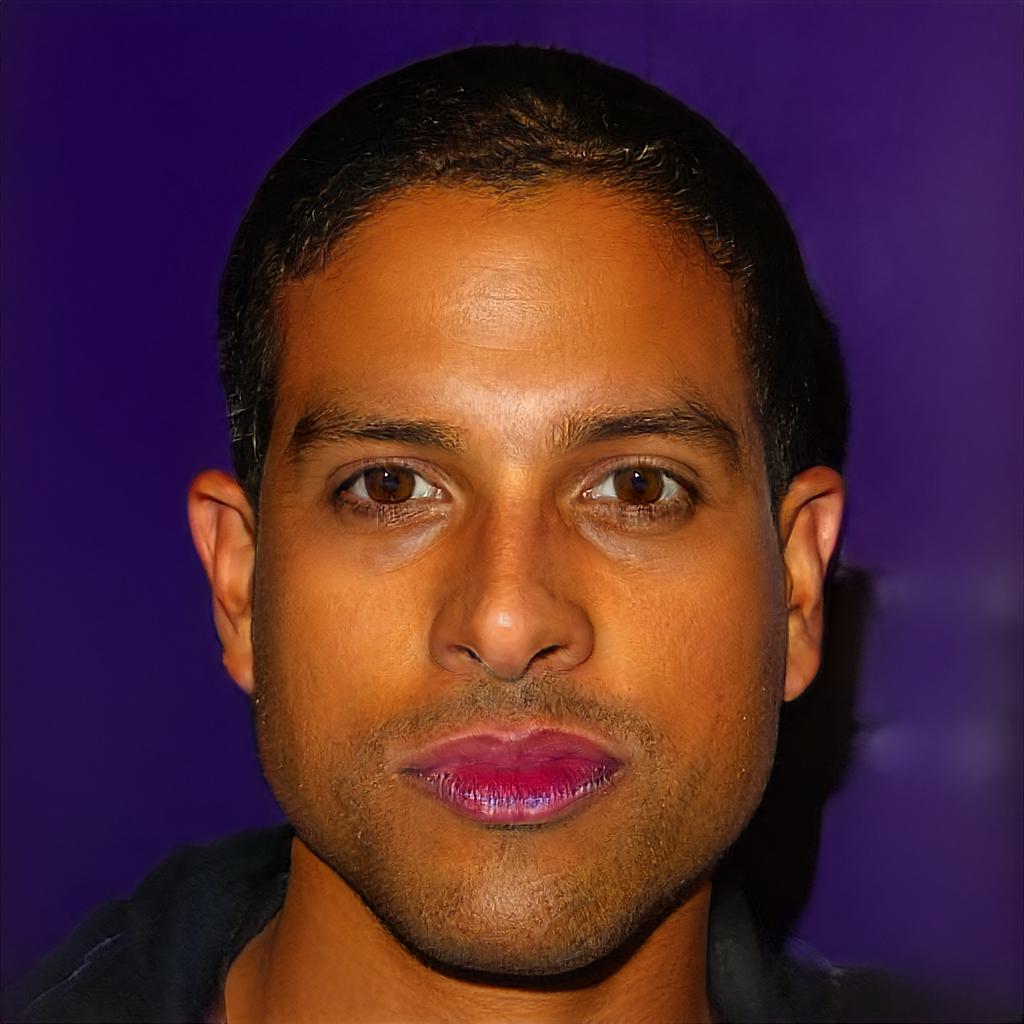}
            \centering
        \end{subfigure}
        \vspace{8pt}
    \end{tabular}}
    \resizebox{\textwidth}{!}
    {\begin{tabular}{ccccccc}
        \begin{subfigure}[t]{0.2\textwidth}
            \includegraphics[width=\linewidth]{images/fig8/085807.jpg}
            \centering
            \vspace{2pt}
            \fontsize{12pt}{14pt}\selectfont{Target 2}
        \end{subfigure}
        \begin{subfigure}[t]{0.2\textwidth}
            \includegraphics[width=\linewidth]{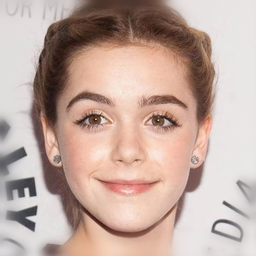}
            \centering
            \vspace{2pt}
            \includegraphics[width=\linewidth]{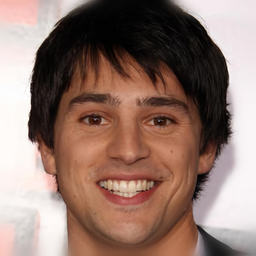}
            \centering
        \end{subfigure}
        \begin{subfigure}[t]{0.2\textwidth}
            \includegraphics[width=\linewidth]{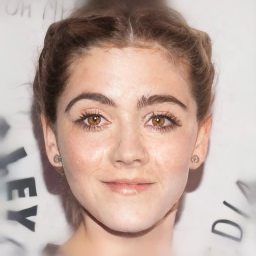}
            \centering
            \vspace{2pt}
            \includegraphics[width=\linewidth]{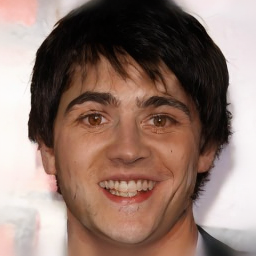}
            \centering
        \end{subfigure}
        \begin{subfigure}[t]{0.2\textwidth}
            \includegraphics[width=\linewidth]{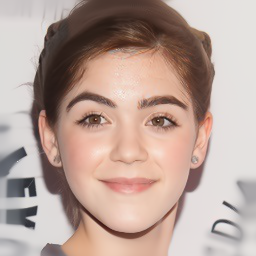}
            \centering
            \vspace{2pt}
            \includegraphics[width=\linewidth]{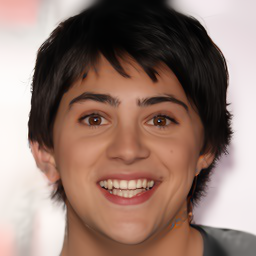}
            \centering
        \end{subfigure}
        \begin{subfigure}[t]{0.2\textwidth}
            \includegraphics[width=\linewidth]{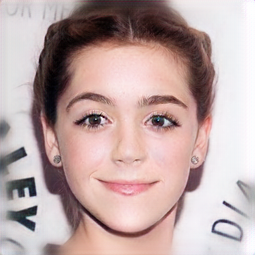}
            \centering
            \vspace{2pt}
            \includegraphics[width=\linewidth]{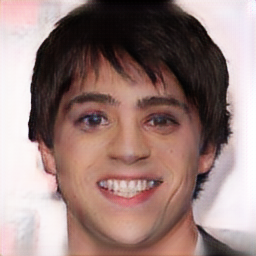}
            \centering
        \end{subfigure}
        \begin{subfigure}[t]{0.2\textwidth}
            \includegraphics[width=\linewidth]{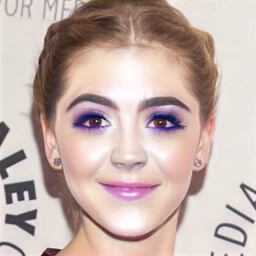}
            \centering
            \vspace{2pt}
            \includegraphics[width=\linewidth]{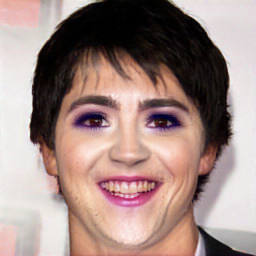}
            \centering
        \end{subfigure}
        \begin{subfigure}[t]{0.2\textwidth}
            \includegraphics[width=\linewidth]{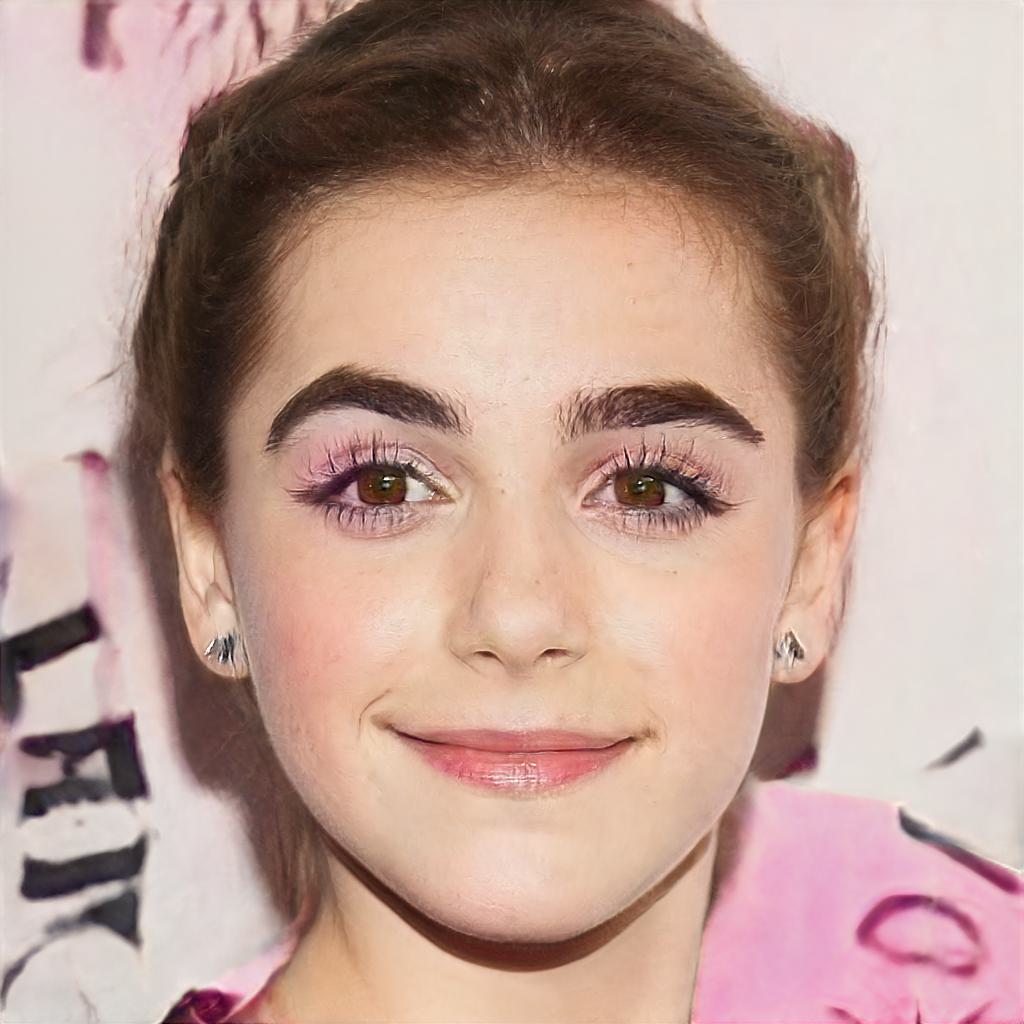}
            \centering
            \vspace{2pt}
            \includegraphics[width=\linewidth]{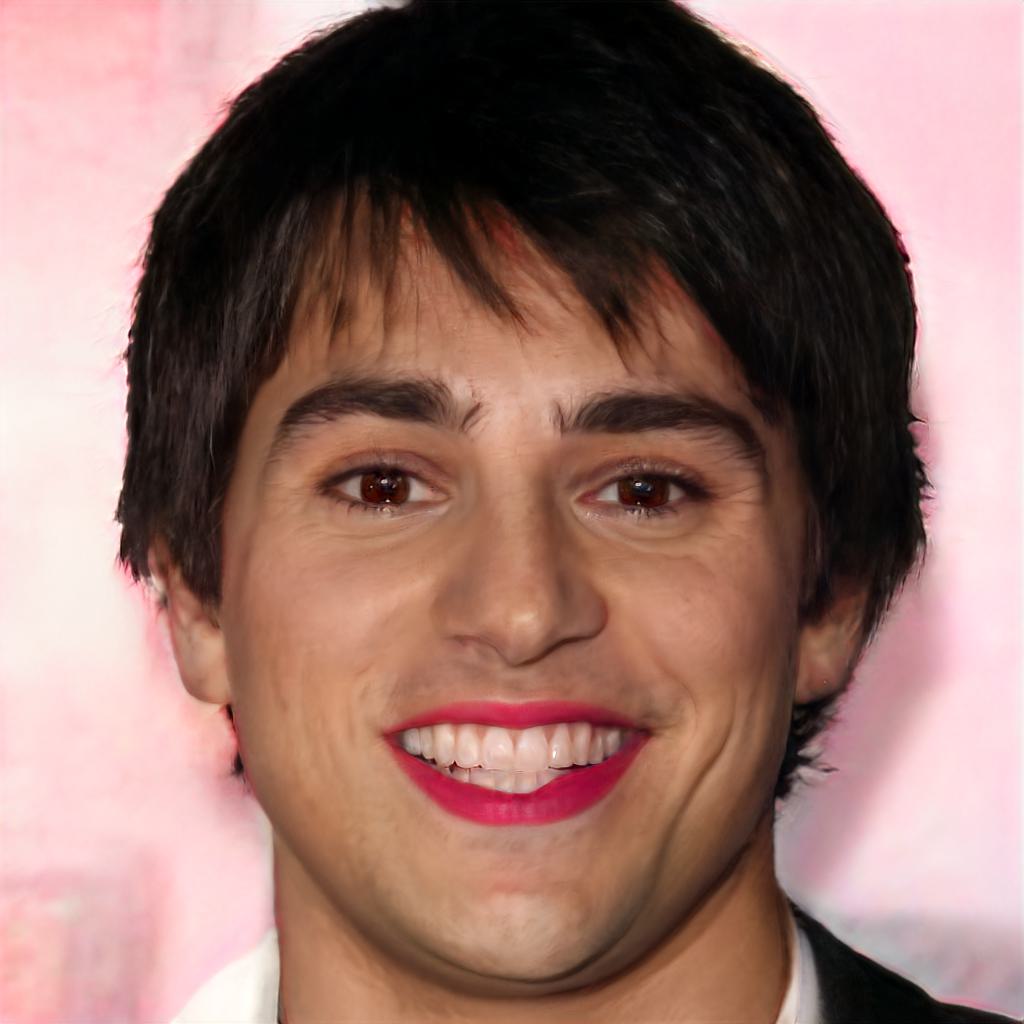}
            \centering
        \end{subfigure}
        \vspace{8pt}
    \end{tabular}}
    \resizebox{\textwidth}{!}
    {\begin{tabular}{ccccccc}
        \begin{subfigure}[t]{0.2\textwidth}
            \includegraphics[width=\linewidth]{images/fig8/116481.jpg}
            \centering
            \vspace{2pt}
            \fontsize{12pt}{14pt}\selectfont{Target 3}
        \end{subfigure}
        \begin{subfigure}[t]{0.2\textwidth}
            \includegraphics[width=\linewidth]{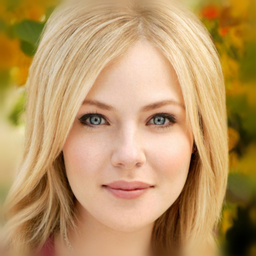}
            \centering
            \vspace{2pt}
            \includegraphics[width=\linewidth]{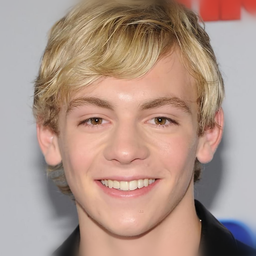}
            \centering
        \end{subfigure}
        \begin{subfigure}[t]{0.2\textwidth}
            \includegraphics[width=\linewidth]{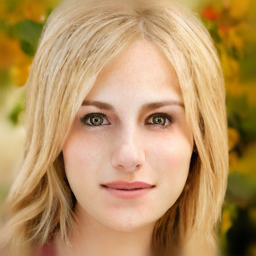}
            \centering
            \vspace{2pt}
            \includegraphics[width=\linewidth]{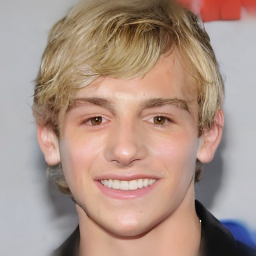}
            \centering
        \end{subfigure}
        \begin{subfigure}[t]{0.2\textwidth}
            \includegraphics[width=\linewidth]{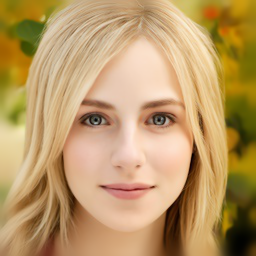}
            \centering
            \vspace{2pt}
            \includegraphics[width=\linewidth]{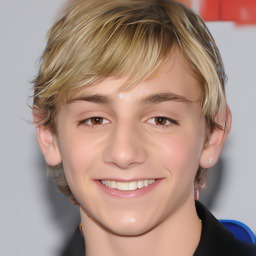}
            \centering
        \end{subfigure}
        \begin{subfigure}[t]{0.2\textwidth}
            \includegraphics[width=\linewidth]{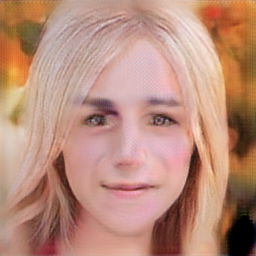}
            \centering
            \vspace{2pt}
            \includegraphics[width=\linewidth]{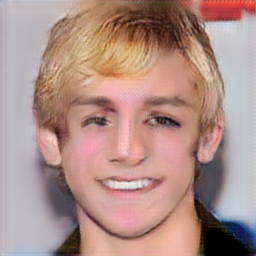}
            \centering
        \end{subfigure}
        \begin{subfigure}[t]{0.2\textwidth}
            \includegraphics[width=\linewidth]{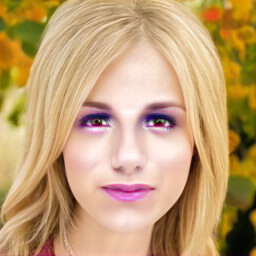}
            \centering
            \vspace{2pt}
            \includegraphics[width=\linewidth]{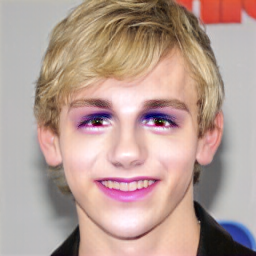}
            \centering
        \end{subfigure}
        \begin{subfigure}[t]{0.2\textwidth}
            \includegraphics[width=\linewidth]{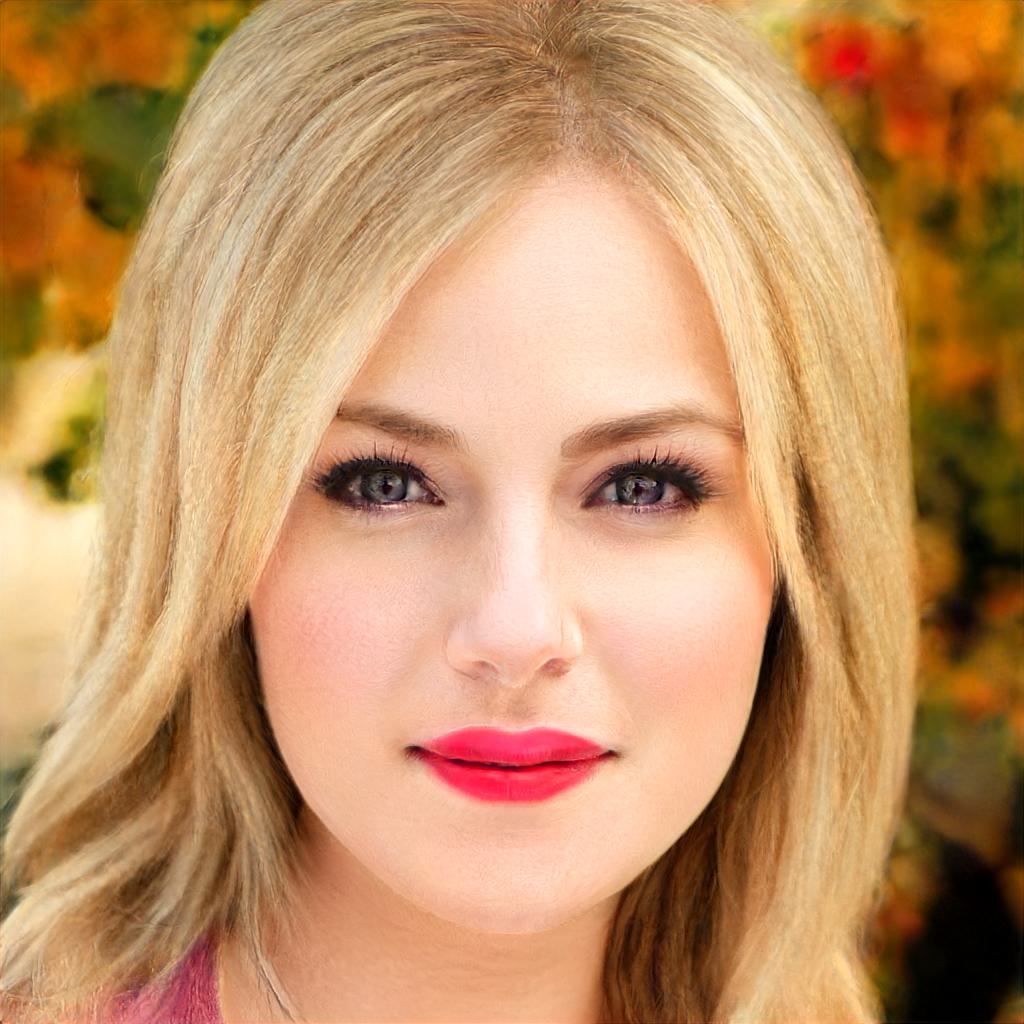}
            \centering
            \vspace{2pt}
            \includegraphics[width=\linewidth]{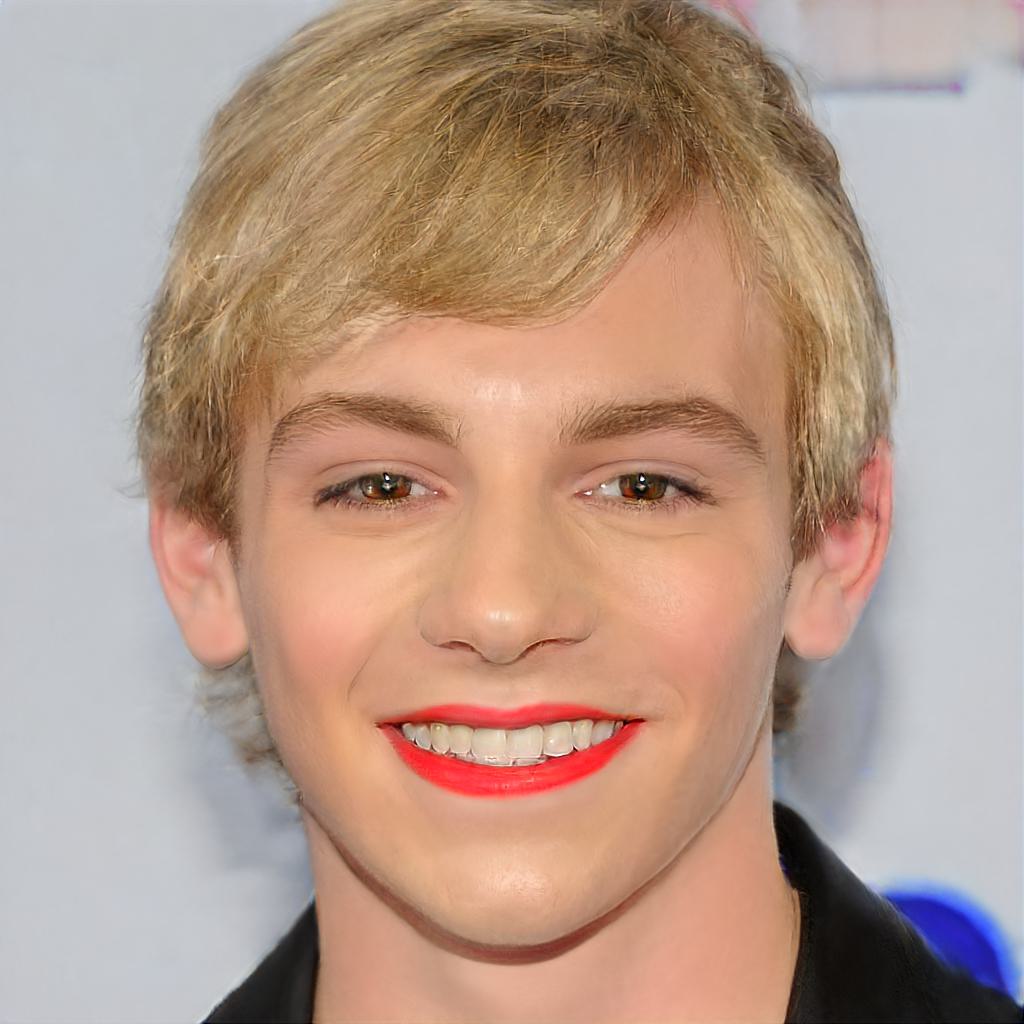}
            \centering
        \end{subfigure}
        \vspace{8pt}
    \end{tabular}}
    \resizebox{\textwidth}{!}
    {\begin{tabular}{ccccccc}
        \begin{subfigure}[t]{0.2\textwidth}
            \includegraphics[width=\linewidth]{images/fig8/169284.jpg}
            \centering
            \vspace{2pt}
            \fontsize{12pt}{14pt}\selectfont{Target 4}
        \end{subfigure}
        \begin{subfigure}[t]{0.2\textwidth}
            \includegraphics[width=\linewidth]{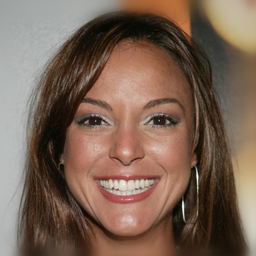}
            \centering
            \vspace{2pt}
            \includegraphics[width=\linewidth]{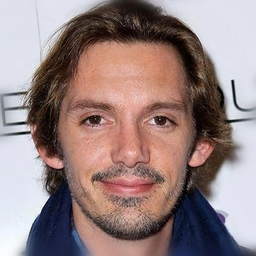}
            \centering
            \vspace{2pt}
            \fontsize{12pt}{14pt}\selectfont{Original Images}
        \end{subfigure}
        \begin{subfigure}[t]{0.2\textwidth}
            \includegraphics[width=\linewidth]{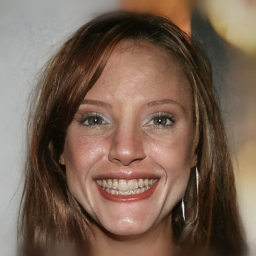}
            \centering
            \vspace{2pt}
            \includegraphics[width=\linewidth]{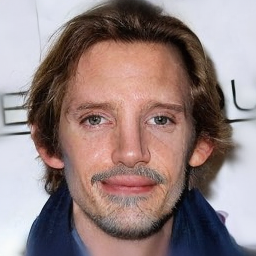}
            \centering
            \vspace{2pt}
            \fontsize{12pt}{14pt}\selectfont{\textbf{Ours}}
        \end{subfigure}
        \begin{subfigure}[t]{0.2\textwidth}
            \includegraphics[width=\linewidth]{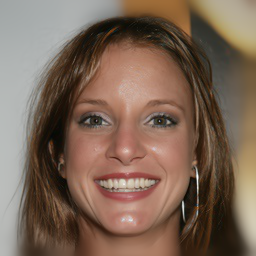}
            \centering
            \vspace{2pt}
            \includegraphics[width=\linewidth]{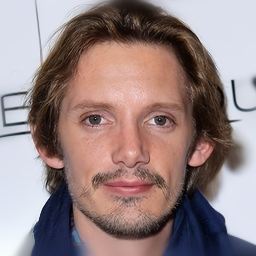}
            \centering
            \vspace{2pt}
            \fontsize{12pt}{14pt}\selectfont{DiffProtect \cite{Diffprotect_2023_arXiv}}
        \end{subfigure}
        \begin{subfigure}[t]{0.2\textwidth}
            \includegraphics[width=\linewidth]{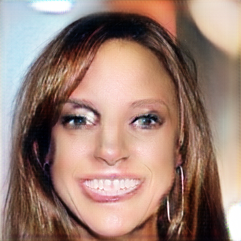}
            \centering
            \vspace{2pt}
            \includegraphics[width=\linewidth]{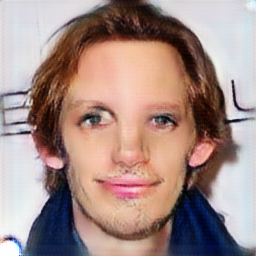}
            \centering
            \vspace{2pt}
            \fontsize{12pt}{14pt}\selectfont{AMT-GAN \cite{AMT-GAN_2022_CVPR}}
        \end{subfigure}
        \begin{subfigure}[t]{0.2\textwidth}
            \includegraphics[width=\linewidth]{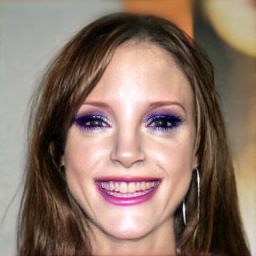}
            \centering
            \vspace{2pt}
            \includegraphics[width=\linewidth]{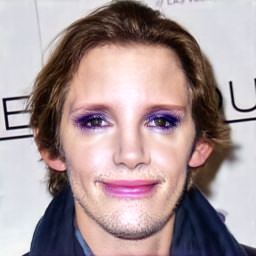}
            \centering
            \vspace{2pt}
            \fontsize{12pt}{14pt}\selectfont{DiffAM \cite{DiffAM_2024_CVPR}}
        \end{subfigure}
        \begin{subfigure}[t]{0.2\textwidth}
            \includegraphics[width=\linewidth]{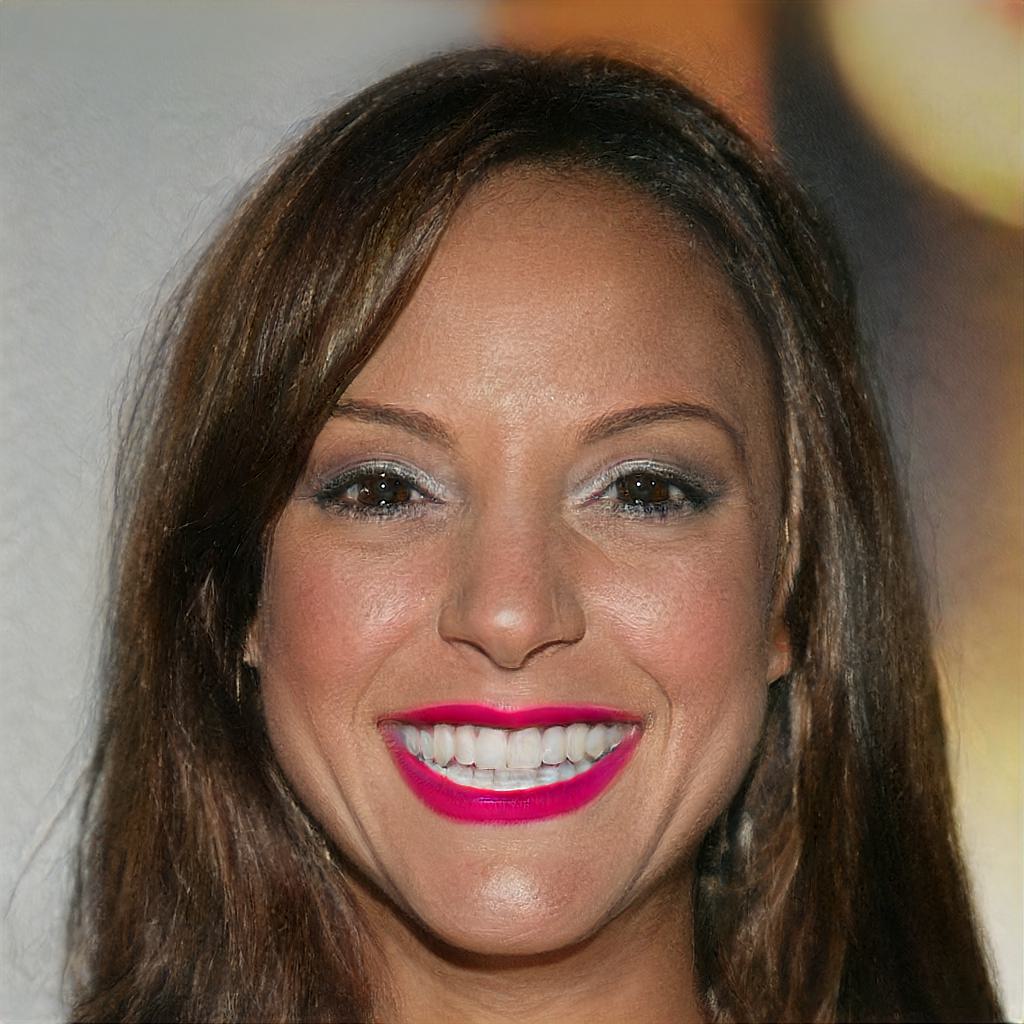}
            \centering
            \vspace{2pt}
            \includegraphics[width=\linewidth]{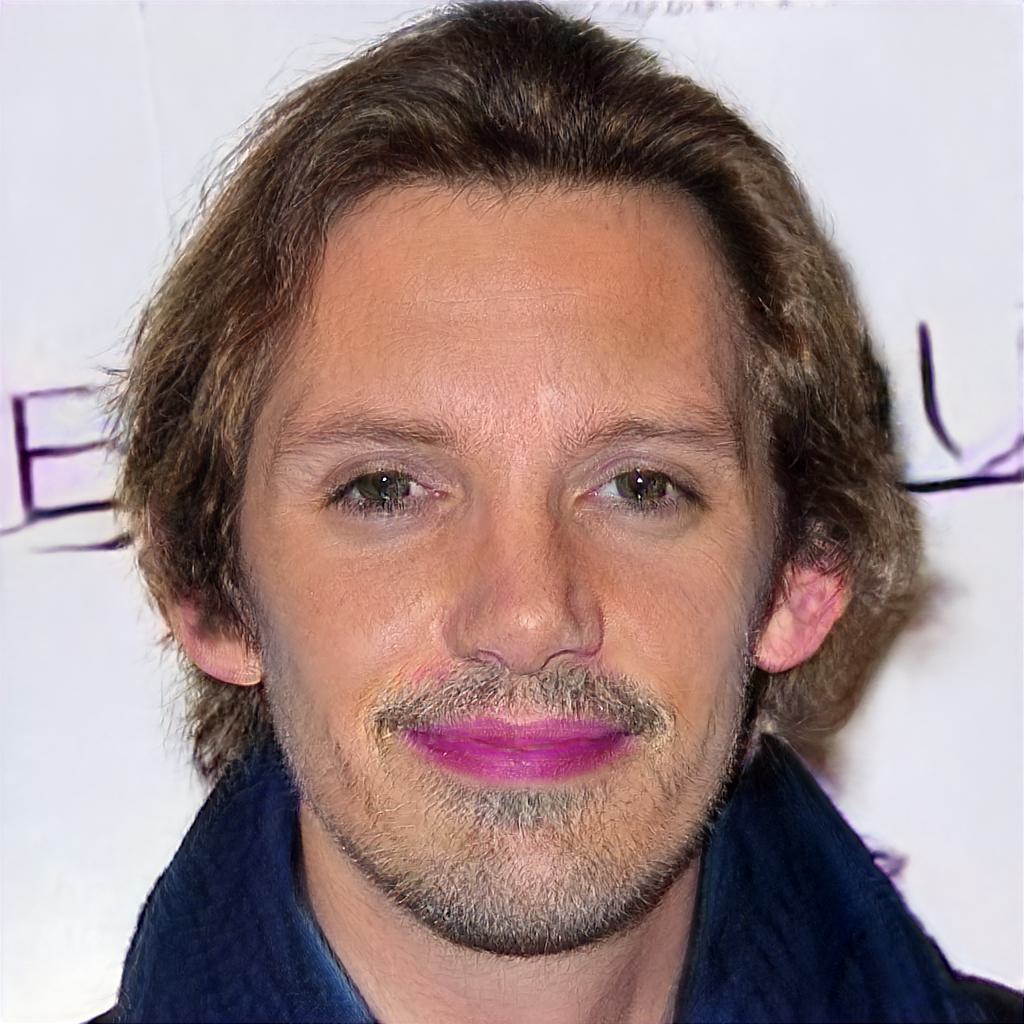}
            \centering
            \vspace{2pt}
            \fontsize{12pt}{14pt}\selectfont{CLIP2Protect \cite{Clip2protect_2023_CVPR}}
        \end{subfigure}
    \end{tabular}}
    \caption{Visual assessment of the protected images generated by previous methods and our approach for impersonation. Target images for each group are shown on the left side. Original images are selected from the CelebA-HQ \cite{CelebA-HQ_2017_arXiv} dataset.}
    \label{fig_11}
\end{figure*}

\begin{figure*}[h]
    \centering
    \resizebox{\textwidth}{!}
    {\begin{tabular}{ccccccc}
        \begin{subfigure}[t]{0.2\textwidth}
            \includegraphics[width=\linewidth]{images/fig7/fake.png}
            \vspace{2pt}
            \centering
            \fontsize{12pt}{14pt}\selectfont{Synthesized target}
        \end{subfigure}
        \begin{subfigure}[t]{0.2\textwidth}
            \includegraphics[width=\linewidth]{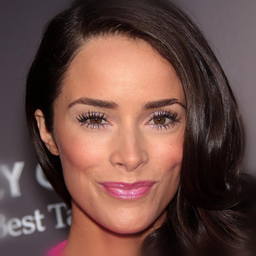}
            \centering
            \vspace{2pt}
            \includegraphics[width=\linewidth]{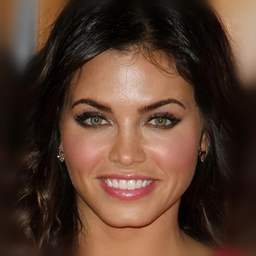}
            \centering
            \vspace{2pt}
            \includegraphics[width=\linewidth]{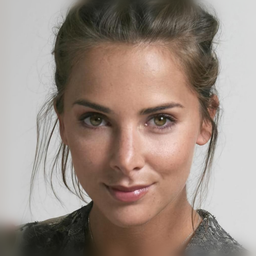}
            \centering
            \vspace{2pt}
            \fontsize{12pt}{14pt}\selectfont{(a) Original Images}
        \end{subfigure}
        \begin{subfigure}[t]{0.2\textwidth}
            \includegraphics[width=\linewidth]{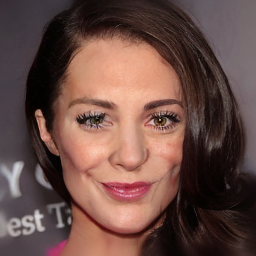}
            \centering
            \vspace{2pt}
            \includegraphics[width=\linewidth]{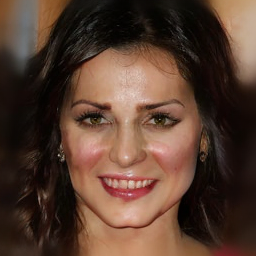}
            \centering
            \vspace{2pt}
            \includegraphics[width=\linewidth]{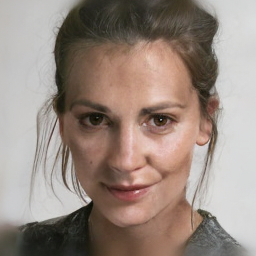}
            \centering
            \vspace{2pt}
            \fontsize{12pt}{14pt}\selectfont{(b) Impersonation}
        \end{subfigure}
        \begin{subfigure}[t]{0.2\textwidth}
            \includegraphics[width=\linewidth]{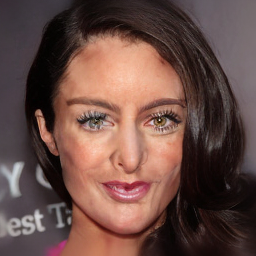}
            \centering
            \vspace{2pt}
            \includegraphics[width=\linewidth]{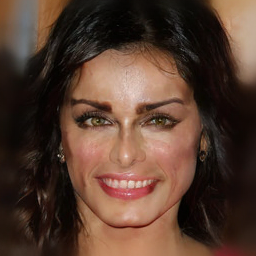}
            \centering
            \vspace{2pt}
            \includegraphics[width=\linewidth]{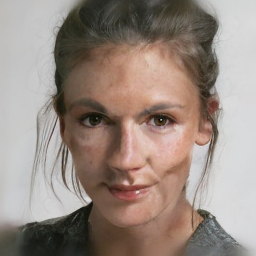}
            \centering
            \vspace{2pt}
            \fontsize{12pt}{14pt}\selectfont{(c) Obfuscation}
        \end{subfigure}
        \begin{subfigure}[t]{0.2\textwidth}
            \includegraphics[width=\linewidth]{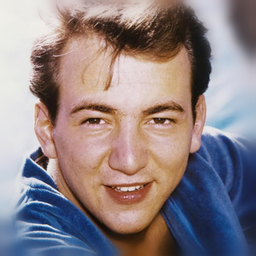}
            \centering
            \vspace{2pt}
            \includegraphics[width=\linewidth]{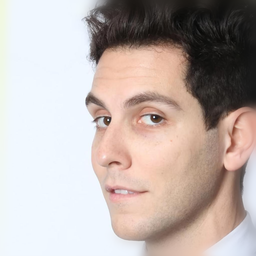}
            \centering
            \vspace{2pt}
            \includegraphics[width=\linewidth]{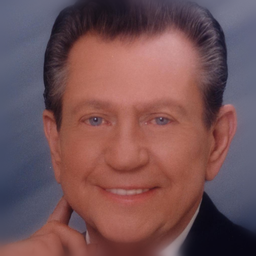}
            \centering
            \vspace{2pt}
            \fontsize{12pt}{14pt}\selectfont{(d) Original Images}
        \end{subfigure}
        \begin{subfigure}[t]{0.2\textwidth}
            \includegraphics[width=\linewidth]{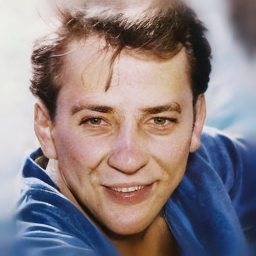}
            \centering
            \vspace{2pt}
            \includegraphics[width=\linewidth]{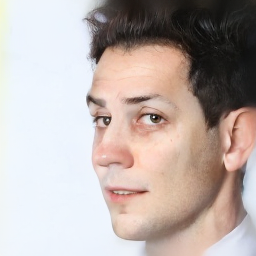}
            \centering
            \vspace{2pt}
            \includegraphics[width=\linewidth]{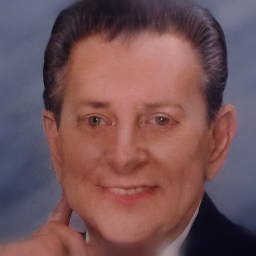}
            \centering
            \vspace{2pt}
            \fontsize{12pt}{14pt}\selectfont{(e) Impersonation}
        \end{subfigure}
        \begin{subfigure}[t]{0.2\textwidth}
            \includegraphics[width=\linewidth]{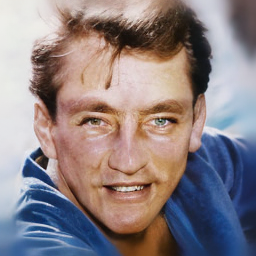}
            \centering
            \vspace{2pt}
            \includegraphics[width=\linewidth]{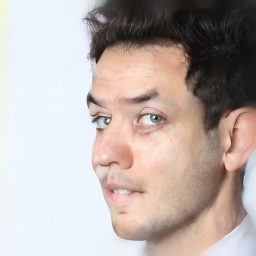}
            \centering
            \vspace{2pt}
            \includegraphics[width=\linewidth]{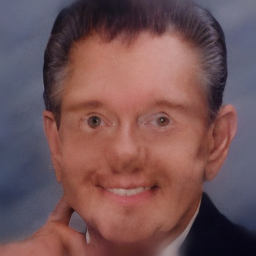}
            \centering
            \vspace{2pt}
            \fontsize{12pt}{14pt}\selectfont{(f) Obfuscation}
        \end{subfigure}
        \vspace{8pt}
    \end{tabular}}
    \caption{Visual assessment of the protected images generated by both impersonation and obfuscation losses and those generated with only the obfuscation loss. The synthesized target image is shown on the left side. (a) and (d) show original images, which are selected from the CelebA-HQ \cite{CelebA-HQ_2017_arXiv} dataset. (b) and (e) show protected images generated with both impersonation and obfuscation losses. (c) and (f) show protected images generated with only obfuscation loss.}
    \label{fig_12}
\end{figure*}

\textbf{Impersonation.} To show the effectiveness of our proposed method in impersonating different identities, we visually compare the protected face images generated by ours and recent methods in Fig. \ref{fig_11}. Compared to makeup-based methods, i.e., AMT-GAN \cite{AMT-GAN_2022_CVPR}, DiffAM \cite{DiffAM_2024_CVPR} and CLIP2Protect \cite{Clip2protect_2023_CVPR}, which change the makeup styles of the input images and intensify makeup in special parts of the face, our method can better preserve image styles. Compared to DiffProtect \cite{Diffprotect_2023_arXiv}, which changes the facial expressions of the input images and smooths them out, ours preserves facial and hair details and adds perturbation only to identity-related features.\\

\textbf{Obfuscation.} A visual comparison between images generated using a combination of impersonation and obfuscation loss functions and those generated solely with the obfuscation loss function is shown in Fig. \ref{fig_12}. The results demonstrate that the images generated with both losses simultaneously appear more natural and exhibit fewer distortions. This suggests that incorporating an impersonation objective with obfuscation enhances the visual quality of the generated images, producing faces that maintain more realistic features and preserve coherence in appearance.
\section{Limitations and Future Directions}
Given an input and target image, our approach generates the protected image in approximately 15 seconds on average, outperforming DiffProtect \cite{Diffprotect_2023_arXiv} ($\approx$19 seconds) and CLIP2Protect \cite{Clip2protect_2023_CVPR} ($\approx$30 seconds). All experiments were conducted on a single Nvidia GeForce RTX 4090. Despite its faster performance, the protection time of our approach can be further reduced by leveraging multiple GPUs and parallel computing optimizations. While AMT-GAN \cite{AMT-GAN_2022_CVPR} and DiffAM \cite{DiffAM_2024_CVPR} generate protected images in under one second, they require re-training the entire model for each new target identity, making them less flexible in practical scenarios.\\
In future work, we plan to replace the current surrogate model-based training paradigm, which involves iterative image reconstruction during latent code optimization, with a more efficient attack strategy that operates directly within the semantic space of the UNet proposed by An \etal. \cite{SD4Privacy_2024_ICME}. This shift is expected to accelerate the execution time of our method significantly.

\end{document}